\documentclass[sigplan,nonacm]{acmart}
\usepackage{xcolor}
\usepackage{gensymb}
\usepackage{soul}
\usepackage{siunitx}
\usepackage{enumitem}
\usepackage{makecell}
\usepackage{subcaption} 
\usepackage{algorithm}
\usepackage{multirow}
\usepackage{algpseudocode}

\newcommand{\circled}[1]{\textcircled{\scriptsize #1}}
\begin{document}

\title{Scalable Disk-Based Approximate Nearest Neighbor Search with Page-Aligned Graph
}

\author{Dingyi Kang$^\alpha$, Dongming Jiang$^\alpha$, Hanshen Yang$^\beta$, Hang Liu$^\beta$ and Bingzhe Li$^\alpha$ \\
$^\alpha$University of Texas at Dallas\\
$^\beta$Rutgers University
}

\begin{abstract}
Approximate Nearest Neighbor Search (ANNS), as the core of vector databases (VectorDBs), has become widely used in modern AI and ML systems, powering applications from information retrieval to bio-informatics. While graph-based ANNS methods achieve high query efficiency, their scalability is constrained by the available host memory. Recent disk-based ANNS approaches mitigate memory usage by offloading data to Solid-State Drives (SSDs). However, they still suffer from issues such as long I/O traversal path, misalignment with storage I/O granularity, and high in-memory indexing overhead, leading to significant I/O latency and ultimately limiting scalability for large-scale vector search.

In this paper, we propose PageANN, a disk-based approximate nearest neighbor search (ANNS) framework designed for high performance and scalability. PageANN introduces a page-node graph structure that aligns logical graph nodes with physical SSD pages, thereby shortening I/O traversal paths and reducing I/O operations. Specifically, similar vectors are clustered into page nodes, and a co-designed disk data layout leverages this structure with a merging technique to store only representative vectors and topology information, avoiding unnecessary reads. To further improve efficiency, we design a memory management strategy that combines lightweight indexing with coordinated memory–disk data allocation, maximizing host memory utilization while minimizing query latency and storage overhead. Experimental results show that PageANN significantly outperforms state-of-the-art (SOTA) disk-based ANNS methods, achieving 1.85x–10.83x higher throughput and 51.7\%–91.9\% lower latency across different datasets and memory budgets, while maintaining comparable high recall accuracy.
\end{abstract}


\maketitle
\pagestyle{plain}
\section{Introduction}
Approximate Nearest Neighbor Search (ANNS), as the core of vector databases (VectorDBs), is a transformative technology that enables the extraction of meaning from massive datasets. ANNS is widely applied in domains such as machine learning ~\cite{cao2017binary, bijalwan2014knn, tagami2017annexml, cover1967nearest, zhu2019accelerating, flickner1995query, zhang2022uni, zhang2018visual, huang2020embedding, lewis2020retrieval, guu2020retrieval, borgeaud2022improving, asai2023retrieval, liu2023learning, ram2023context, izacard2023atlas} and bioinformatics ~\cite{bittremieux2018fast, schutze2022nearest}, where the ability to efficiently identify data points most similar to a query drive key advancements. 
Recently, graph-based ANNS approaches~\cite{malkov2014approximate, hnsw16, chen2018sptag, fu2017fast, nsg19, li2019approximate} have gained popularity due to their high query efficiency, achieved through fast graph traversal strategies. However, as vector datasets scale up (e.g., billions of vectors~\cite{fu2017fast, li2018design, wei2020analyticdb}), these methods face scalability challenges, since both the vectors and their associated indexes may not be fully stored in host memory.

To address this scalability challenge in large-scale vector search, disk-based graph ANNS solutions have been developed~\cite{diskann, diskann++, lm-diskann, starling, spann, pipeann}. These methods offload the data to disk and load it into memory only when needed, thereby reducing memory consumption. For example, the pioneer work DiskANN~\cite{diskann} store both the full vectors and the graph index on SSDs while retaining compressed vector representations in memory. During search, only the original vectors and the neighbor information along the search path are loaded from disk into memory. Subsequent studies such as  SPANN ~\cite{spann}, Starting ~\cite{starling}, and PipeANN ~\cite{pipeann} adopt similar designs to further improve query efficiency for large-scale ANN search. However, these methods still face significant I/O latency, particularly under constrained host memory. Moreover, as datasets continue to scale, such approaches encounter persistent scalability bottlenecks. 

\begin{figure}[!t]
    \centering
\includegraphics[width=0.95\linewidth]{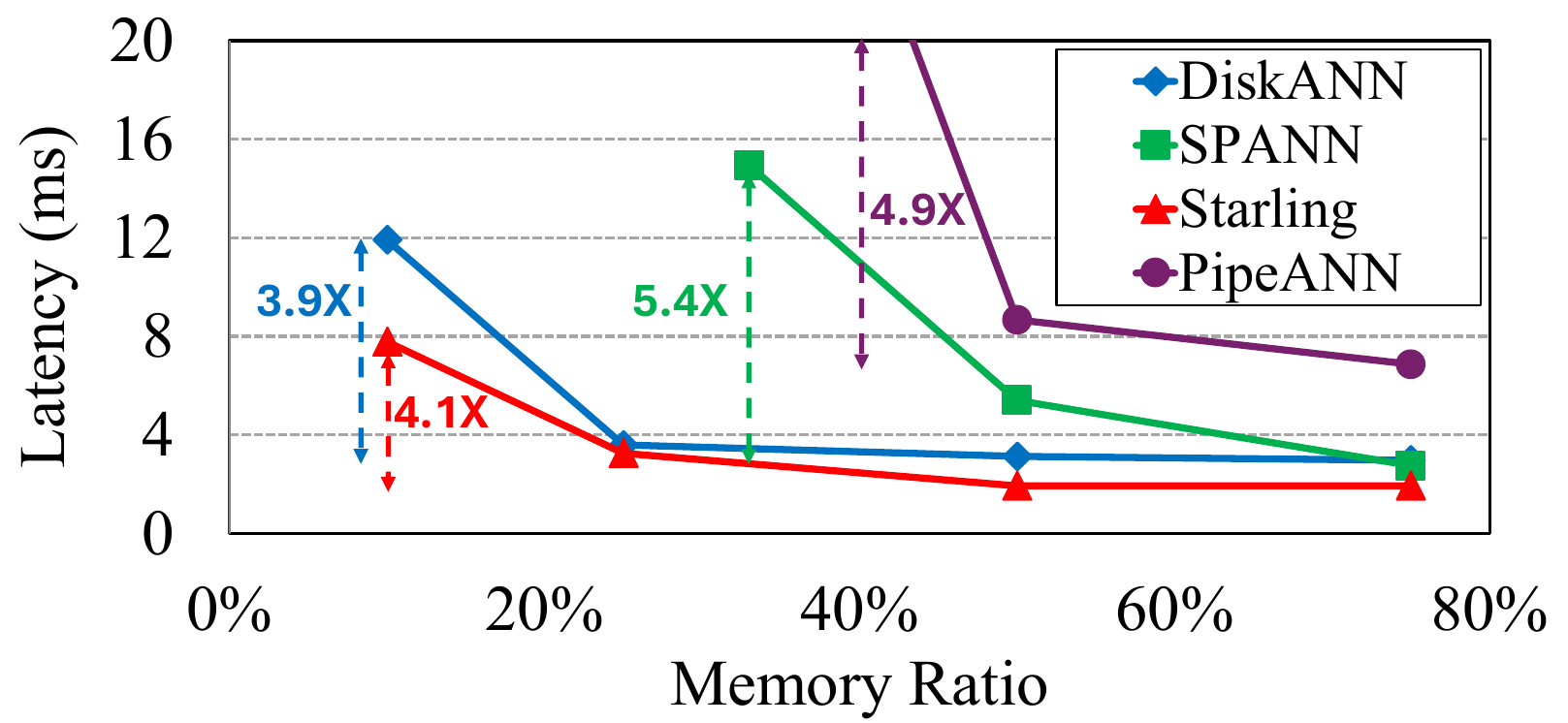} 
        \caption{Impact of memory ratio on query latency for state-of-the-art disk-based ANN schemes (i.e., DiskANN~\cite{diskann}, SPANN~\cite{spann}, Starling~\cite{starling}, and PipeANN~\cite{pipeann}) on the SIFT dataset. The memory ratio represents the proportion of the dataset size that is available in host memory. (Due to the minimum memory requirement, PipeANN and SPANN cannot run with the memory ratio below 30\%.)}
\label{fig:motivation_scalability}
\end{figure}
As shown in Figure~\ref{fig:motivation_scalability}, existing schemes achieve relatively low query latencies when sufficient memory is available. However, as the available memory decreases, latencies increase significantly. For example, reducing the memory ratio from 50\% to 10\% can result in more than a threefold increase in latency. In some cases, such as SPANN and PipeANN, a minimum memory threshold is required for execution (e.g., SPANN cannot operate with a memory ratio below 30\%). Moreover, with higher memory ratio, the performance of those schemes are saturated. These observations underscore the scalability limitations of current disk-based ANN schemes as the size of vector datasets continues to grow.

To analyze the limitations of state-of-the-art (SOTA) disk-based ANN schemes, we identify three primary issues: \textbf{(1) Long I/O traversal path:} As the number of vectors grows, the corresponding graph expands significantly. Although graph algorithms such as Vamana~\cite{diskann} can reduce graph diameter and the average number of query hops, fewer search hops do not necessarily translate to reduced I/Os. \textbf{(2) Misalignment with storage I/O granularity:} During the search process, each step typically requires reading only a small amount of data (e.g., one or a few vectors per SSD page). However, SSDs must read entire pages to fulfill these requests. The misalignment between graph node size and SSD I/O granularity leads to wasted I/O bandwidth and underutilized reads. \textbf{(3) High in-memory indexing overhead:} Although SOTA disk-based ANN methods reduce memory usage by offloading part of the search data to disk, they still reserve substantial memory for essential data structures (e.g., compressed vector data for distance calculation or graph index for fast routing). For large-scale datasets, this memory requirement remains prohibitively high, limiting both applicability and scalability.

Given these limitations, we aim to enhance the scalability of disk-based ANN systems by developing a new vector indexing structure and reconstructing the data layout in storage. However, several critical questions must be addressed: \textit{(1) What kind of graph structure can be designed to shorten the graph traversal paths and reduce I/O operations?} The gap between theoretical graph hops and the actual number of I/O operations must be bridged to improve overall query performance. \textit{(2) How can the vector data layout in storage be reconstructed to align with I/O granularity?} While some prior work addresses vector locality by grouping them into a single SSD page, further improvements require jointly considering graph topology and vector data placement to optimize I/O performance. \textit{(3) What in-memory indexing structure can balance both performance and scalability?} Existing methods often incur substantial memory overhead for storing index data or compressed vectors to accelerate queries. Therefore, a new indexing structure must be co-designed with the system to simultaneously achieve high performance and scalability.

To address these challenges, we propose PageANN, a framework that enables efficient large-scale ANN search under various memory budgets while maintaining high recall accuracy. The key innovation is a page-node graph structure for query search, which differs from prior vector-based graphs that treat each vector as an individual node in graphs. The page-node graph bridges the gap between theoretical hop counts and the number of read I/Os by aligning each logical page node with a physical SSD page. To construct this graph, we design a new algorithm that groups similar vectors into logical page nodes and leverages only representative vectors to establish connections with neighboring pages, ensuring both accuracy and efficiency.

Second, we design a new SSD data layout that cooperates with the page-node graph to optimize query processing. Each logical page node in the graph is allocated to a physical SSD page, and only representative vectors from neighboring pages are selected, to form inter-page connections. We incorporate a merging mechanism within each physical page to remove redundant data, resulting in storing more data and accelerating the search process. Additionally, each SSD page embeds its own topology information (i.e., neighbor IDs and their compressed values), eliminating the need for extra reads during query execution.

Lastly, we propose a new lightweight index that quickly routes incoming queries to page nodes near the target, reducing the length of the search path. Moreover, a memory-disk coordination strategy that determines search-related data stored in disk or memory based on the available memory budget. This combined approach improves the performance of disk-based ANN systems while maintaining low memory overhead and high scalability.

In summary, we develop the PageANN framework, comprising around 6K lines of code, and release it as open source\footnote{https://github.com/Dingyi-Kang/PageANN}. Compared to other SOTA schemes, PageANN achieves from 1.85x to 10.83x higher throughput and reduces latency by 51.7\% to 91.9\% across datasets and memory budgets, while maintaining the same high recall accuracy.

The remainder of this paper is organized as follows: Section~\ref{sec:background} presents the background and motivation for ANN. Section~\ref{sec:motivation} discuss the specific motivation for this work. Section~\ref{sec:design} introduces the the overall architecture of PageANN and its algorithms. Section~\ref{sec:implementation} details its implementation. Section~\ref{sec:results} shows the experimental results. Related work is given in Section~\ref{sec:related}. Finally, conclusions are drawn in Section~\ref{sec:conclusion}.

\section{Background}\label{sec:background}

In this section, we present the background and basics of nearest neighbor search, and introduce SOTA disk-based graph ANNS schemes.

\subsection{Nearest Neighbor Search}
In the nearest neighbor search (NNS) problem in high dimensional spaces, data points are represented as \(d\)-dimensional vectors in \(\mathbb{R}^d\), and the Euclidean distance is typically used as the distance metric. Throughout this paper, we use the terms \textit{points} and \textit{vectors} interchangeably. The Euclidean distance between two data points \(p\) and \(q\) is denoted as \(\|p - q\|_2\).

Given a dataset \(X = \{x_1, x_2, \dots, x_n\}\), the \(k\)-nearest neighbor (\(k\)NN) search for a query point \(q\) is defined as finding a subset \(k\text{NN}(q) \subseteq X\) such that:
\begin{itemize}
    \item \(|k\text{NN}(q)| = k\), and
    \item \(\forall x \in k\text{NN}(q), \forall x' \in X \setminus k\text{NN}(q), \|x - q\|_2 \leq \|x' - q\|_2\).
\end{itemize}

However, in high-dimensional spaces, the computational cost of exact nearest neighbor search becomes prohibitively expensive due to the curse of dimensionality~\cite{lsh98}. Consequently, research has increasingly focused on approximate nearest neighbor search (ANNS) methods, which trade a small amount of accuracy for substantial gains in efficiency.

Let the result returned by an ANNS algorithm be \(R\), where \(|R| = k\). The quality of \(R\) is commonly evaluated using recall@k, defined as:
\[
\text{Recall@k} = \frac{|R \cap k\text{NN}(q)|}{k}.
\]

\subsection{Graph-based ANNS}
Unlike tree-based methods~\cite{bentley1975multidimensional, fukunaga2006branch, silpa2008optimised} and hashing-based methods~\cite{lsh98, datar2004locality, tao2009quality, gan2012locality, huang2015query, liu2014sk, lu2020r2lsh, lu2020vhp, tian2023db}, which suffer from severe performance degradation and scalability issues as datasets grow, graph-based ANNS approaches~\cite{malkov2014approximate, hnsw16, chen2018sptag, fu2017fast, nsg19, li2019approximate} have emerged as one of the most effective solutions for large-scale search in high-dimensional spaces, providing both high efficiency and accuracy.

In graph-based ANNS, vector data are represented as nodes connected to their nearest neighbors under a distance metric, forming a sparse yet well-connected topology that enables efficient navigation. The search begins from one or more entry points and iteratively progresses toward closer neighbors until a stopping condition is met, after which the top-$k$ closest nodes are returned. By limiting distance computations to small local neighborhoods at each step, graph-based ANNS greatly reduces computational overhead. Moreover, by traversing proximity-based connections between data points, these methods effectively alleviate the curse of dimensionality and avoid boundary issues that commonly affect partitioning- and hashing-based approaches.


Although graph-based ANNS methods achieve excellent speed and accuracy, they often incur scalability issue due to substantial memory overhead. To address this issue, a range of disk-based graph ANNS solutions has been proposed~\cite{diskann, diskann++, lm-diskann, starling, spann, pipeann}, which reduce memory consumption by offloading data to disk while striving to maintain high search performance. For example, DiskANN\cite{diskann}, Starling\cite{starling}, and PipeANN\cite{pipeann} store original vectors and graph topology on SSDs while retaining only product quantization (PQ)–compressed vectors in memory. In contrast, SPANN\cite{spann} keeps the graph index entirely in memory and offloads only the vector data to SSDs, performing graph traversal in memory and issuing I/O requests only after traversal completes.
\begin{figure}[!t] 
        \centering
        \includegraphics[width=0.95\linewidth]{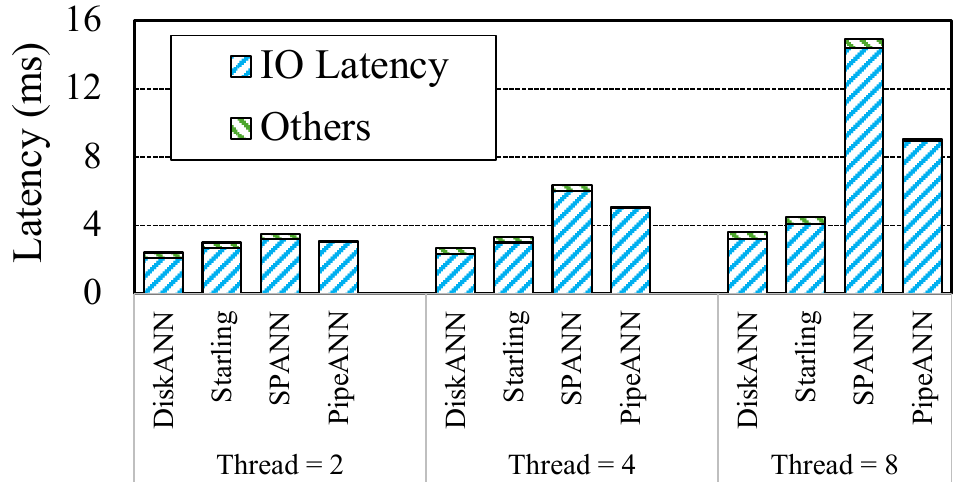} 
        \caption{Query latency breakdown analysis for SOTA schemes.}
        \label{fig:motivation_IO}
\end{figure}

\begin{figure*}[!t]
    \centering
\includegraphics[width=0.9\linewidth]{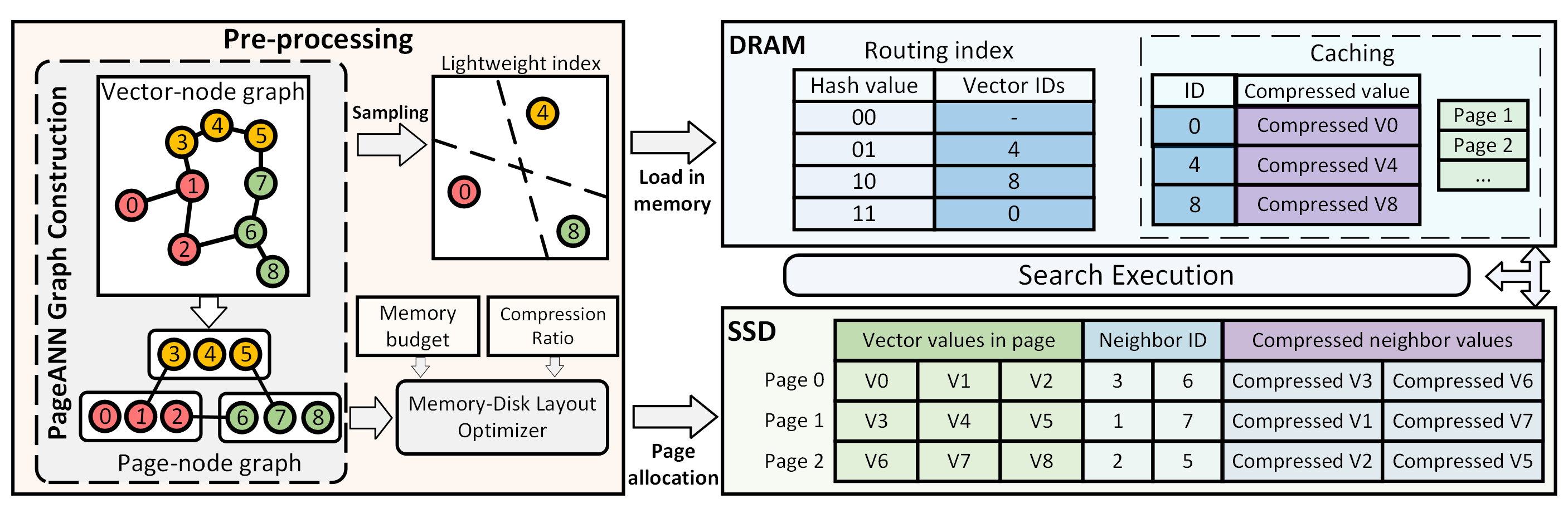} 
        \caption{Overall structure of PageANN.}
\label{fig:overall}
\end{figure*}
\section{Motivation}\label{sec:motivation}

Although existing disk-based solutions alleviate some issues in graph-based ANNS, they still suffer from severe search latency and scalability limitations. As shown in Figure~\ref{fig:motivation_scalability}, SOTA disk-based ANN experience scalability issues, particularly under limited host memory, where query latency can increase more than threefold compared to a sufficient memory configuration (e.g., 40\% memory ratio). Furthermore, our analysis of performance bottlenecks, illustrated in Figure~\ref{fig:motivation_IO}, shows that I/O latency accounts for over 90\% of the total query latency across all schemes. Based on these observations, we identify the following reasons for the low scalability and performance of the SOTA schemes.


\textbf{Long I/O graph traversal path.} SOTA disk-based graph ANNS methods suffer from high I/O latency, primarily due to long traversal paths on disk during query processing. Each hop along the path triggers a batch of I/O read requests, which accumulate into substantial overhead as the search progresses. With increasing dataset size, the average length of search path grows, leading to more disk seeks and cache misses. As a result, query performance is constrained by I/O bottlenecks rather than computation, posing a significant challenge for large-scale deployments. Although Vamana graph construction~\cite{diskann} aims to reduce graph diameter (i.e., the maximum number of hops between any two nodes), such reductions in theoretical hops do not necessarily translate into fewer I/O operations. In fact, suboptimal data layouts can further prolong I/O traversal paths due to a mismatch between logical graph hops and actual storage accesses.


\begin{table}[!t]
\centering
\small
\caption{Read amplification of SOTA Disk-based ANN schemes.}
\label{tab:read_amplification}
\resizebox{0.9\columnwidth}{!}{%
\begin{tabular}{lccc}
\toprule
Scheme\textbackslash
Dataset & SIFT100M & SPACEV100M & DEEP100M \\
\midrule
DiskANN  & 18.29 & 20.08 & 8.00 \\
Starling & 1.71 & 1.95 & 1.28 \\
PipeANN  & 18.29 & 20.08 & 8.00 \\
SPANN    & 2.0 & 2.0 & 2.0 \\
\bottomrule
\end{tabular}%
} 
\end{table}

\textbf{Misalignment of storage IO granularity.} 
Another major source of excessive I/O latency is the mismatch between storage I/O granularity and the size of individual vectors. Even if the theoretical minimum graph size reduces the upper bound on I/O operations (i.e., the number of hops equals to the number of I/Os), the data in SSD pages may not be fully utilized due to misalignment between page granularity and graph node size, resulting in suboptimal latency. Modern SSDs typically enforce a minimum read unit of one page (often 4 KB or larger), which is significantly larger than a single vector (e.g., 128 bytes in the SIFT dataset~\cite{sift1B}). Prior work has attempted to mitigate this problem by clustering similar nodes into the same page to reduce read amplification. However, the irregular structure of graphs and the unpredictable nature of query paths make it difficult to ensure that all data within a page are relevant and useful. Consequently, state-of-the-art schemes inevitably incur wasted reads during query processing. For example, as shown in Table~\ref{tab:read_amplification}, on the SIFT dataset~\cite{sift1B} with an average graph degree of 24, existing methods exhibit read amplification from 1.28 to 20.8 during search. Therefore, efficiently addressing this misalignment is critical for eliminating read amplification and further reducing I/O cost.

\textbf{Large in-memory indexing overhead.} 
Although state-of-the-art disk-based schemes claim to offload data to storage to improve the scalability of ANN systems, they still require a substantial amount of memory for compressed data or indexing structures. For example, SPANN reserves more than 30\% of the dataset size in memory for graph indexing. Below this threshold, it cannot operate reliably. DiskANN and Starling store compressed vector data in memory to accelerate queries. While they can operate with as little as 10\% of the dataset size in memory, this comes with trade-offs: either reduced accuracy due to lossy compression or significantly increased latency caused by longer traversal paths needed to maintain accuracy. In summary, despite their claims of scalability through disk offloading, current disk-based systems remain constrained by substantial memory requirements, resulting in high latency or limited scalability.

\section{PageANN Design}\label{sec:design}
In this section, we present the design of PageANN with a page-node graph structure that coordinates the logical graph with the physical characteristics of storage to accelerate query search and improve system scalability. As illustrated in Figure~\ref{fig:overall}, PageANN operates in two major stages: pre-processing and query processing.
In the pre-processing stage, vectors are organized into a page-node graph. Based on the available memory budget and a predefined compression ratio, the system determines the amount of compressed data to be allocated in memory. A lightweight index is also constructed to enable efficient query routing.
In the query processing stage, the combination of caching and lightweight indexing enables fast traversal of the graph. A new disk layout is employed to support the page-node graph structure and to reduce I/O operations for subsequent disk-resident graph retrievals. The following subsections detail the specific design considerations for each component.
\begin{figure}[!t]
    \centering
\includegraphics[width=0.99\linewidth]{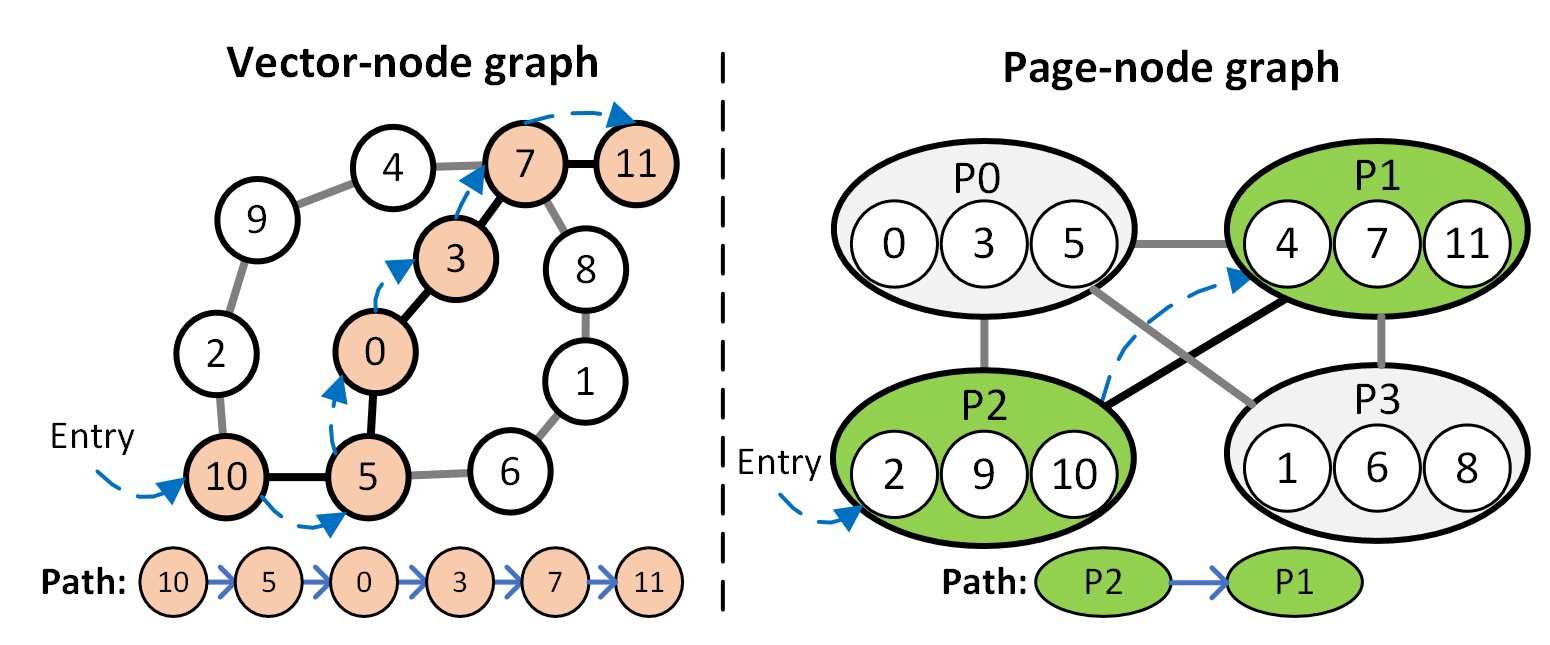} 
        \caption{Main idea of page-node graph.}
\label{fig:page-graph}
\end{figure}
\subsection{Page-Node Graph Construction in PageANN}\label{sec:graphconstruction}
The main idea of page-node graph in PageANN is to shift the granularity of the index graph from individual vectors to page nodes, aligning the graph structure with the I/O granularity of storage systems. As shown in Figure~\ref{fig:page-graph}, this page-node graph preserves essential relationship between vectors while substantially reducing the overall graph size. However, designing such a graph introduces two key challenges.
\textit{(1) Translating reduced traversal paths in the graph into reduced I/O.} Shorter traversal paths in the graph do not necessarily lead to fewer I/O operations if the graph structure is not coordinated with storage characteristics. Consequently, the design must consider both the logical properties of the graph and the constraints of SSD hardware, as well as the intended search process. \textit{(2) How should a page node be represented in the graph?} Each page node contains multiple vectors, and constructing meaningful neighbor connections is non-trivial. Using too few representative vectors to represent a page node (e.g., a single proxy vector for all vectors in a page) risks missing true neighbors due to poor coverage of the intra-page distribution. Using too many representatives can lead to redundant edges, increasing both search cost and storage overhead.

\begin{algorithm}[!t]
\caption{Page-Node Graph Construction}
\label{alg:page_graph_construction}
\begin{algorithmic}[1]
\Require Vector-based graph $G_0$; page-node capacity $n$; hop parameter $h$ for candidate selection;
\Ensure Page-node graph stored on SSD, where each page contains its vectors and neighbors

\State $P \gets$ empty list of pages
\State \textit{UngroupedVectors} $\gets$ all vectors in $G_0$
\While{\textit{UngroupedVectors} $\neq \emptyset$}
    \State $v \gets \textit{UngroupedVectors.extract()}$; $p \gets [v]$
    \State $C \gets G_0.\textit{ungroupedNbrsWithinHops}(v, h)$
    \State $V \gets$ top $(n-1)$ closest vectors to $v$ from $C$
    \State $p.\textit{append}(V)$
    \State $\textit{UngroupedVectors.remove}(V)$
    \While{$|p| < n$ and \textit{UngroupedVectors} $\neq \emptyset$}
        \State $p.\textit{append}(\textit{UngroupedVectors.extract()})$
    \EndWhile
    \State $P.\textit{append}(p)$
\EndWhile
\ForAll{page $p \in P$}
    \State \textit{PageVecs} $\gets$ empty list
    \State \textit{PageNbrs} $\gets$ empty set
    \ForAll{vector $v \in p$}
        \State \textit{PageVecs.append}($G_0.\textit{getValue}(v)$)
        \ForAll{neighbor $u \in G_0.\textit{getNbrs}(v)$}
            \If{$u \notin p$}
                \State \textit{PageNbrs.insert}($u$)
            \EndIf
        \EndFor
    \EndFor
    \State \textit{writeIntoSSDPage}(\textit{PageVecs}, \textit{PageNbrs})
\EndFor
\end{algorithmic}
\end{algorithm}

To address these challenges, we propose a topology-guided construction scheme that derives the page-node graph from a high-quality vector-based graph, as shown in Algorithm~\ref{alg:page_graph_construction}. We start with the Vamana graph~\cite{diskann}, known for producing low-diameter, high-quality graphs, though our method is modular and can use any disk-friendly graph construction algorithm. As illustrated in Lines 1-13 of Algorithm~\ref{alg:page_graph_construction}, vectors are first grouped into page-nodes by clustering vectors with their $n-1$ nearest neighbors within $h$ hops, where $n$ is the target page-node capacity and $h$ is a tunable parameter. Once grouped, the external neighbors of each page node are determined by aggregating the vector-level edges from all its constituent vectors to vectors in other page nodes, removing intra-node edges, and merging duplicate connections to the same external neighbors (Lines 14-26). This preserves the essential connectivity of the original graph while eliminating redundancy.

The page-node graph offers three main advantages: (1) Aligning graph nodes with physical SSD pages maximizes the utility of each page read, reducing I/O operations during search. This alignment increases the effective search space per I/O read, thereby effectively shortening the search path. (2) Connections are derived from the vector-level graph, preserving the full descriptive power of the original vector relationships without incurring significant space overhead. (3) Merging and avoiding redundant edges (i.e., merging edges connecting to the same neighbors and removing internal connections within one page) reduces storage overhead, enabling more search-relevant data to fit within one page.

\begin{figure}[!t]
    \centering
\includegraphics[width=0.9\linewidth]{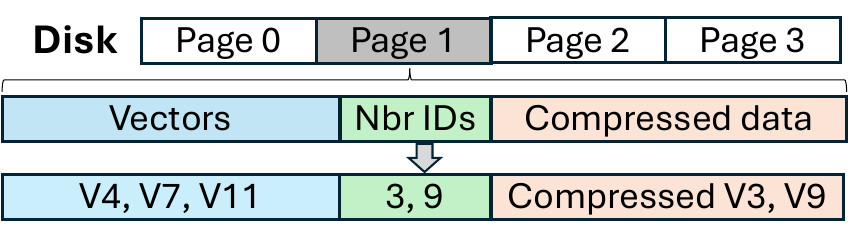} 
        \caption{Disk layout of PageANN based on the graph example in Figure~\ref{fig:page-graph}.}
\label{fig:layout}
\end{figure}
\subsection{A New Disk Layout Supporting Page-node Graph}\label{sec:disklayout}
To fully leverage the benefits of the page-node graph and translate reduced traversal paths into lower I/O overhead, PageANN adopts a new data layout in storage that aligns the page-node abstraction with SSD access characteristics. This design also introduces an additional field that reduces the number of next-hop nodes to fetch, further decreasing I/O requests. As shown in Figure~\ref{fig:layout}, each SSD page contains three major fields: (i) the vector values stored in the page, used for computing the actual distances to the query; (ii) the neighbor IDs, representing the topology of the page-node graph; and (iii) the compressed vectors of page neighbors, used for determining the next hops in the search.

First, as discussed earlier, reducing the number of graph traversal hops alone does not directly translate into fewer I/O requests. To bridge this gap, PageANN maps each page node to exactly one physical SSD page, ensuring that each visited graph node corresponds to a single SSD page read. This mapping allows the entire page data to be fully utilized. In traditional disk-based graph search, the system reads the target node and its neighbor IDs, then issues separate I/O requests to fetch each neighbor for distance computation. In large-scale graphs, where a node may have tens or even hundreds of neighbors, this can result in excessive I/O. To address this, PageANN stores compressed neighbor vectors directly alongside their IDs within the same page, enabling next-hop computations to be performed entirely based on the current page without accessing additional pages.

Moreover, to ensure that all data associated with a page node fits into a single SSD page (typically 4KB, 8KB, or larger), PageANN determines the number of vectors per page node using the following equation:
\[
N_\text{nodes} = 
\frac{
S_\text{page} - 2 \cdot S_\text{num\_nbrs} - S_\text{nbrID} \cdot N_\text{nbrs} + S_\text{CV} \cdot N_\text{CV}
}{
D \cdot S_\text{dtype}
}
\]

where $S_\text{page}$ denotes the page size; $S_\text{num\_nbrs}$ refers to the size of the variable used to store the total number of neighbors; $S_\text{nbrID}$ is the size of each neighbor ID; $N_\text{nbrs}$ is the total number of page neighbors; $S_\text{CV}$ is the size of a compressed vector; $N_\text{CV}$ is the number of neighbors with compressed vectors on disk; $D$ is the vector dimension, and $S_\text{dtype}$ is the size of the vector data type. 

Based on the memory budget and $S_\text{CV}$, we can conservatively estimate the number of compressed values ($N_\text{CV}$) to be stored per page. To save space and accommodate more vectors per page, the system keeps only one copy of each compressed vector across memory and disk. This design not only optimizes SSD page usage but also reduces I/O during search. Further details on memory–disk coordination are provided in the following subsection.




\subsection{PageANN Memory Management}
To improve the scalability of ANNS while maintaining fast query processing, PageANN manages host memory through two components: lightweight indexing and caching with memory–disk cooperation. The lightweight index quickly routes incoming queries to page nodes near the target, reducing the length of the search path. The remaining memory stores compressed vector data and page node, enabling further path determination without accessing storage. Based on the available memory budget, the scheme decides which search-related data to keep in memory, ensuring full utilization of memory resources for optimal performance.

\noindent\textbf{Caching for fast lightweight indexing.} PageANN leverages a memory space to maintain a data caching to support a fast routing procedure. Given a memory budget, PageANN samples a subset of vectors and projects each onto randomly generated hyperplanes defined by a projection matrix. The sign of each projection is encoded as a bit, and the resulting bit sequence forms a binary hash code. Sampled vectors’ IDs are assigned to hash buckets based on these codes. The Hamming distance between hash codes approximates vector similarity, allowing rapid identification of candidates near the query. When additional memory is available, selected page nodes are stored in the caching area. First, PageANN performs a warm-up phase on a dataset to identify the most frequently visited nodes. Within the available budget, PageANN caches the corresponding page nodes for faster retrieval.

This approach offers key advantages: (1) reduced search length by narrowing the search space and providing closer candidates, thereby lowering the average number of hops in the subsequent graph traversal; (2) memory efficiency with lower overhead, since the hash table avoids repeatedly storing the same sampled vector IDs as in graph-based indexing methods.; and (3) fast computation with constant-time generation of query hash codes, bucket retrieval, and vector ID access.

\noindent \textbf{Memory-disk coordination.}
PageANN balances the available memory budget with storage space overhead. As described in Section~\ref{sec:disklayout}, each page node stores compressed neighbor data so that the next step in the search path can be determined directly from the current page. However, embedding these compressed neighbors within a page consumes storage space, reducing the number of vectors that can fit in that page. If this space were instead allocated to vectors, more vectors could be stored per page, further reducing the page-node graph size and thereby improving search efficiency. To leverage this trade-off, PageANN coordinates memory and disk to support the use of high-quality compressed vectors and to maximize the number of vectors stored per page within the given memory budget. This strategy further reduces the size of the page-node graph and amplifies the performance gains from fewer I/O operations.

The memory-disk coordination in PageANN balances accuracy, performance, and storage efficiency according to the available memory budget. (1) Under severely constrained memory conditions, all moderately compressed neighbor values are stored on the same SSD page as their associated page node, with merging and pruning eliminating redundancy and freeing space to accommodate necessary values without significant disk overhead. (2) With moderate memory budgets, PageANN employs a hybrid configuration where some compressed neighbor values are kept in memory while the rest remain on SSDs, enabling higher-accuracy data usage within budget constraints. (3) When memory is sufficient, all compressed neighbor values are stored in memory, and the freed disk space is reallocated to increase the number of vectors per page node, thereby reducing graph size, shortening traversal paths, and minimizing I/O operations to maximize retrieval efficiency.

\subsection{PageANN Graph Search}
Unlike traditional graphs that traverse from vector to vector, PageANN searches by traversing from page to page. To execute the search, PageANN maintains two key data structures: a candidate set, which stores the top-$L$ vectors closest to the query based on compressed representations, and a result set, which holds the IDs of visited vectors along with their full-coordinate distances to the query. The search process unfolds in two phases: an in-memory routing phase followed by an on-disk graph traversal phase as detailed in Algorithm~\ref{alg:search}.
\begin{figure}[!t]
    \centering
\includegraphics[width=0.95\linewidth]{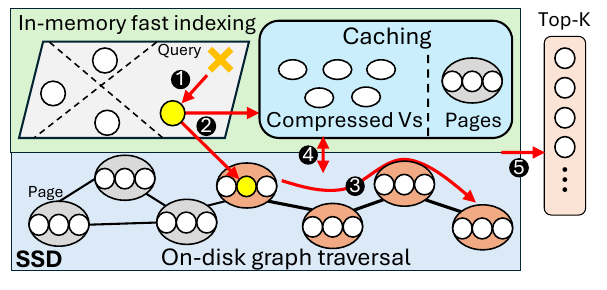}
    \caption{PageANN graph search process.}
    \label{fig:search_process}
\end{figure}

\noindent\textbf{In-memory routing process.} As shown in Figure~\ref{fig:search_process}, upon receiving a query, PageANN computes its binary hash and identifies all hash buckets within a small Hamming radius $r$ (\circled{1}). The vector IDs contained in these buckets are then retrieved, and their approximate distances to the query are estimated using in-memory compressed representations. As shown in Lines 4-7 of Algorithm~\ref{alg:search}, these vector IDs, along with their estimated distances, are inserted into the candidate set, which serves as the set of entry points for subsequent traversal of the page-node graph in disk.

\noindent\textbf{Page-node graph traversal.} 
As shown in Lines 11-19 of Algorithm~\ref{alg:search}, at each hop of the graph traversal, the algorithm selects the closest unvisited vectors from the candidate set, determines their SSD page IDs, and marks them as visited until no unvisited vectors remain in the candidate set. To avoid redundant I/Os, vectors located in the pages that have already been visited or are scheduled for reading are skipped. The collected page IDs are then used to issue batched I/O requests for the corresponding SSD pages.

The target page node will be either read from caching memory or from SSD if it is not in memory (\circled{2}). Using the retrieved data, the next target page node is determined based on the stored topology and the compressed neighbor vectors contained in the SSD page (\circled{3}). During this process, if the required data already reside in memory, PageANN directly accesses and utilizes them without additional retrieval I/O overhead (\circled{4}). Also, the true distances between the query and all vectors in each page are computed and added to the result set along with their vector IDs, while the estimated distances between the query and each page’s neighbor vectors are calculated and added to the candidate set along with the neighbors’ IDs (Lines 20–27 in Algorithm~\ref{alg:search}) (\circled{5}). This expansion process continues iteratively until no unvisited vectors remain in the candidate set. Finally, the result set is sorted by true distance, and the IDs of the top-$k$ closest vectors to the query are returned.

\begin{algorithm}[!t]
\caption{PageANN Graph Search}
\label{alg:search}
\begin{algorithmic}[1]
\Require In-memory hash table $H$, routing search radius $r$, page-node graph $G$, query $q$, I/O batch size $b$
\Ensure Top-$k$ nearest neighbors of $q$
\State $C \gets$ empty heap \hfill {$\triangleright$ Candidate set, a fixed-size min-heap}
\State $V \gets$ empty set \hfill {$\triangleright$ Set of IDs of visited pages}
\State $R \gets$ empty set \hfill {$\triangleright$ Result set}

\State $E \gets H.\textit{getAllVectorIDs}(q, r)$ {$\triangleright$ Retrieve entry vector IDs within Hamming radius $r$}
\ForAll{$v \in E$}
    \State $C \gets C \cup \{(v, \textit{Estimated\_distance}(v, q))\}$
\EndFor

\While{$C$ contains unvisited vectors}
    \State $P \gets$ empty list \hfill {$\triangleright$ Pages to read in this batch}
    \While{$C$ contains unvisited vectors \textbf{and} $|P| < b$}
        \State $v \gets$ closest unvisited vector in $C$
        \State $v.\textit{visited} \gets$ true
        \State $p\_ID \gets \textit{calculate\_pageID}(v)$
        \If{$p\_ID \notin V$}
            \State $V \gets V \cup \{p\_ID\}$
            \State $P \gets P \cup \{p\_ID\}$
        \EndIf
    \EndWhile
    \State $pages\_ready \gets$ \textit{batchReadPagesFromSSD}($P$) 
    \ForAll{page $p \in pages\_ready$}
        \ForAll{$v \in p.\textit{vectors}$}
            \State $R \gets R \cup \{(v, \textit{exact\_distance}(v, q))\}$
        \EndFor
        \ForAll{$n \in p.\textit{neighbors}$}
            \State $C \gets C \cup \{(n, \textit{estimated\_distance}(b, q))\}$
        \EndFor
    \EndFor
\EndWhile
\State $\textit{sortAscendingByDistance}(R)$
\State \Return Top-$k$ vectors from $R$
\end{algorithmic}
\end{algorithm}

\section{\textbf{Implementation}}\label{sec:implementation}
PageANN is implemented in C++ with roughly 6K lines of code and can be deployed under diverse memory budgets, enabling efficient search over large-scale datasets.

\noindent\textbf{Vector ID reassignment and data layout.} 
After constructing the page-node graph, each vector ID is reassigned according to its page-node index and its offset within that page node. A mapping between the original and reassigned vector IDs is maintained to update all neighbor references accordingly during pre-processing stage. Each page node is then written to its corresponding SSD page using the \texttt{std::ofstream} APIs. 

\noindent\textbf{An I/O-computation pipeline.} 
We implement a pipeline that overlaps primary computation with I/O latency by leveraging Linux AIO (Asynchronous I/O) commands such as \texttt{io\_submit} and \texttt{io\_getevents}. A pre-allocated buffer is used to temporarily store vector data from loaded pages. When an asynchronous read completes, the vectors of the corresponding page are placed into the buffer without performing immediate distance computations. Instead, these computations are deferred and executed during the idle wait period of the next pending asynchronous I/O request.

\begin{figure*}[t]
    \centering
    \begin{subfigure}{0.33\textwidth}
        \centering
        \includegraphics[width=\linewidth]{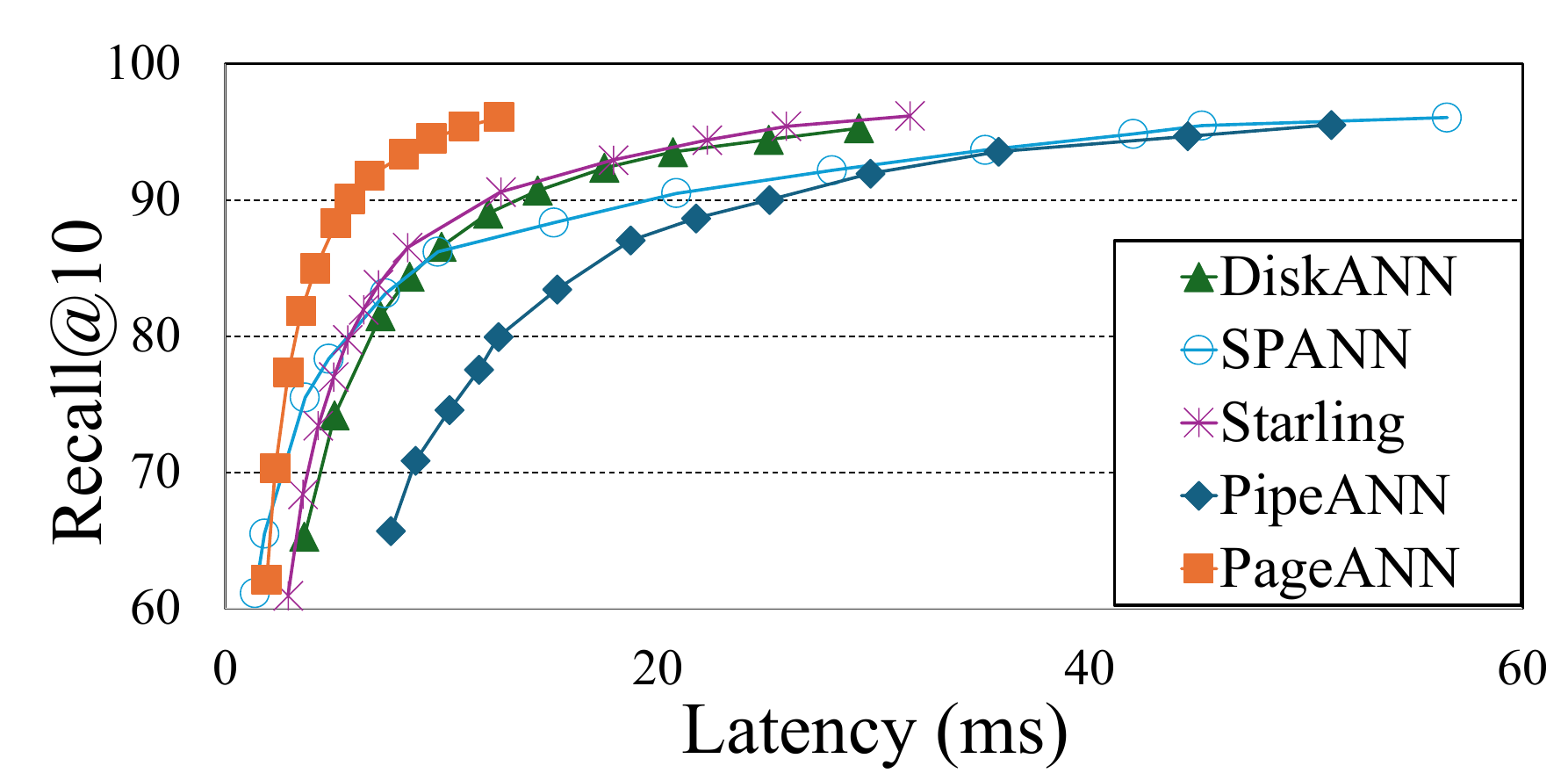} 
        \caption{SIFT100M}
        \label{fig:sift1b_recall_throughput}
    \end{subfigure}
    \hfill
    \begin{subfigure}{0.33\textwidth}
        \centering
        \includegraphics[width=\linewidth]{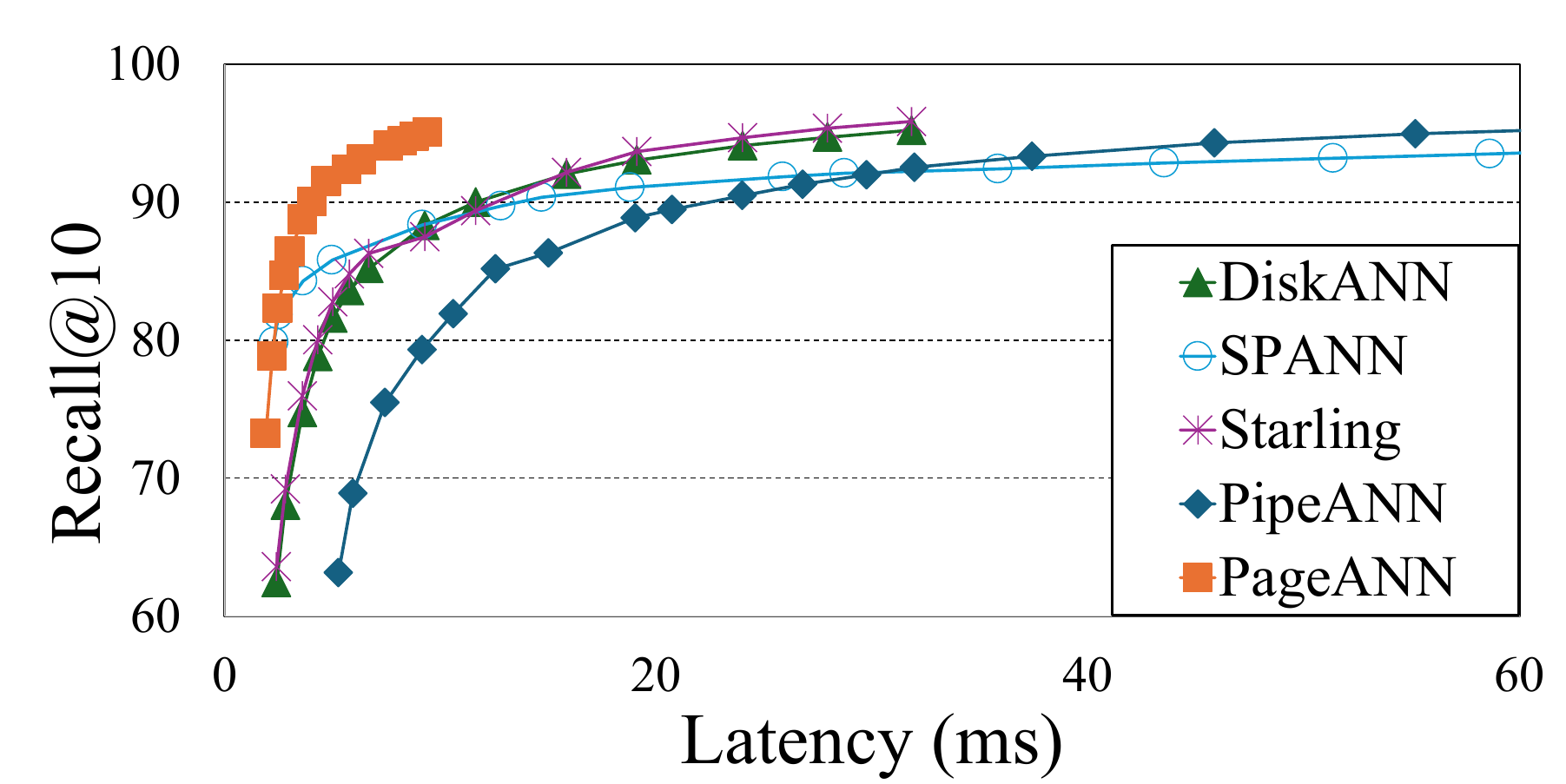} 
        \caption{SPACEV100M}
        \label{fig:spacev100m_recall_latency}
    \end{subfigure}
    \hfill
    \begin{subfigure}{0.33\textwidth}
        \centering
        \includegraphics[width=\linewidth]{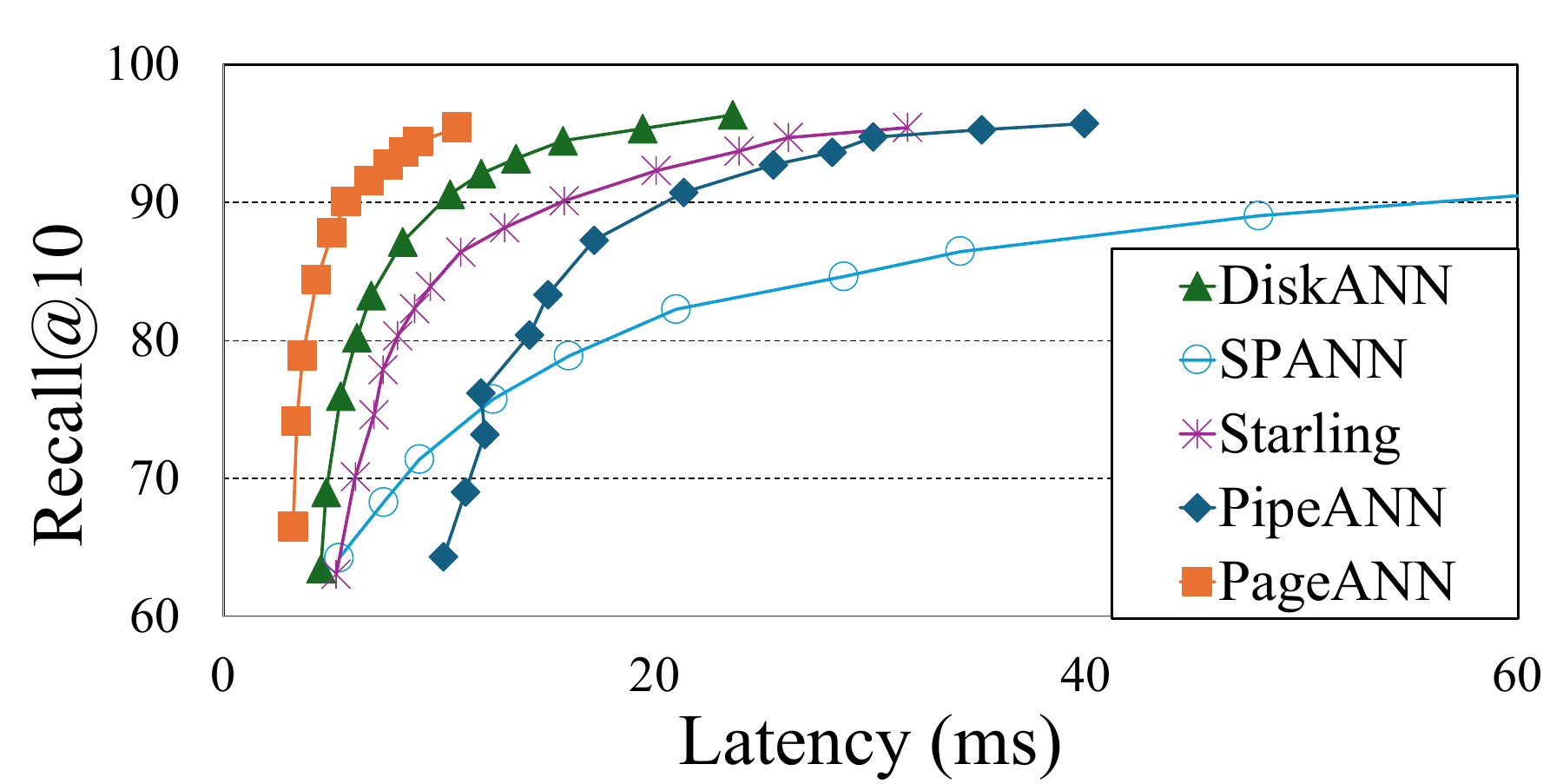} 
        \caption{DEEP100M}
        \label{fig:SPACEV1B_recall_throughput}
    \end{subfigure}
\caption{The query latency vs. Recall@10 with the memory ratio of 30\% for all schemes on million-scale datasets.}

\label{fig:latency_100m}
\end{figure*}

\begin{figure*}[t]
    \centering
    \begin{subfigure}{0.33\textwidth}
        \centering
        \includegraphics[width=\linewidth]{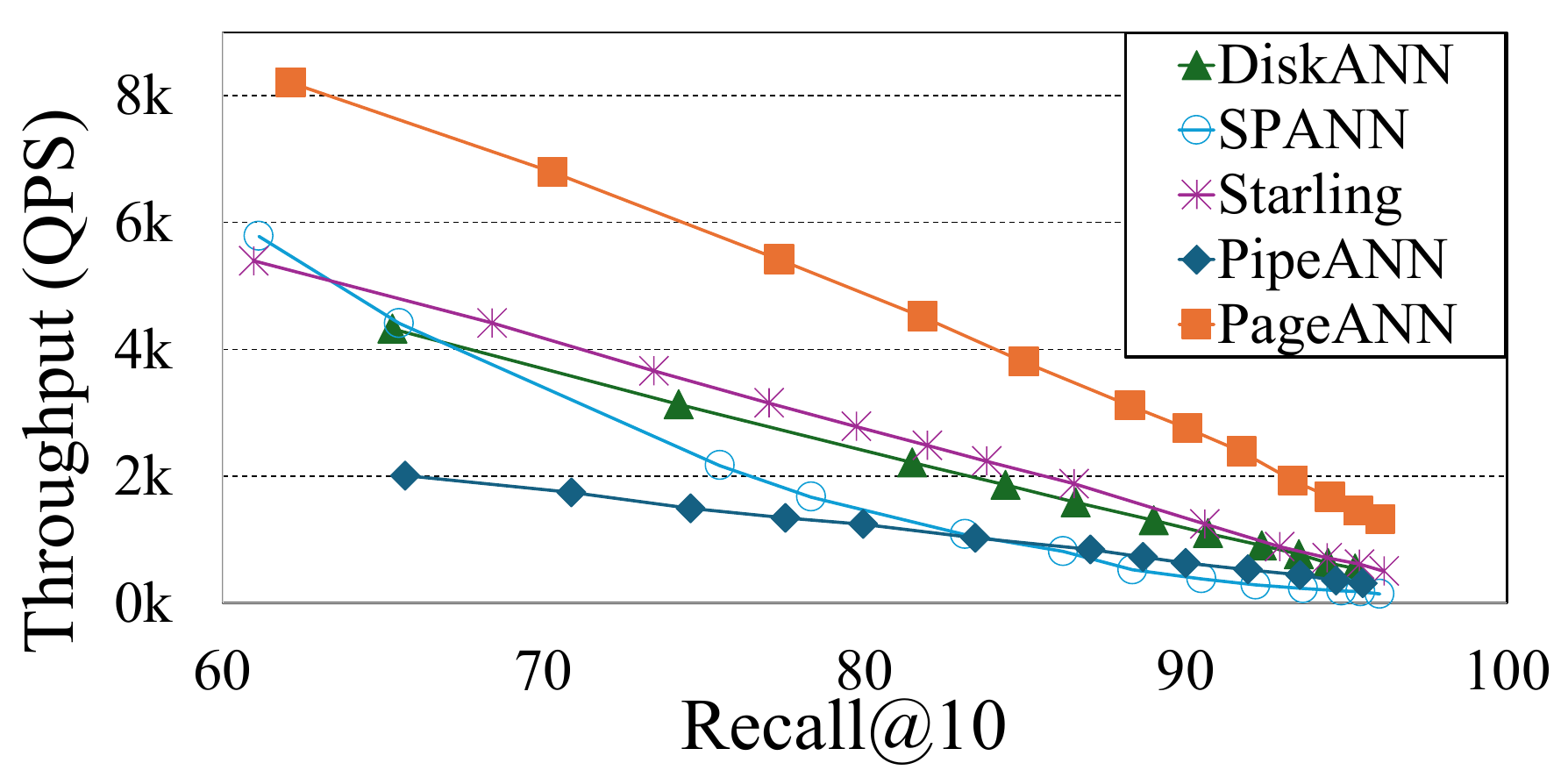}
        \caption{SIFT100M}
        \label{fig:siftthroughput_100m}
    \end{subfigure}
    \hfill
    \begin{subfigure}{0.33\textwidth}
        \centering
        \includegraphics[width=\linewidth]{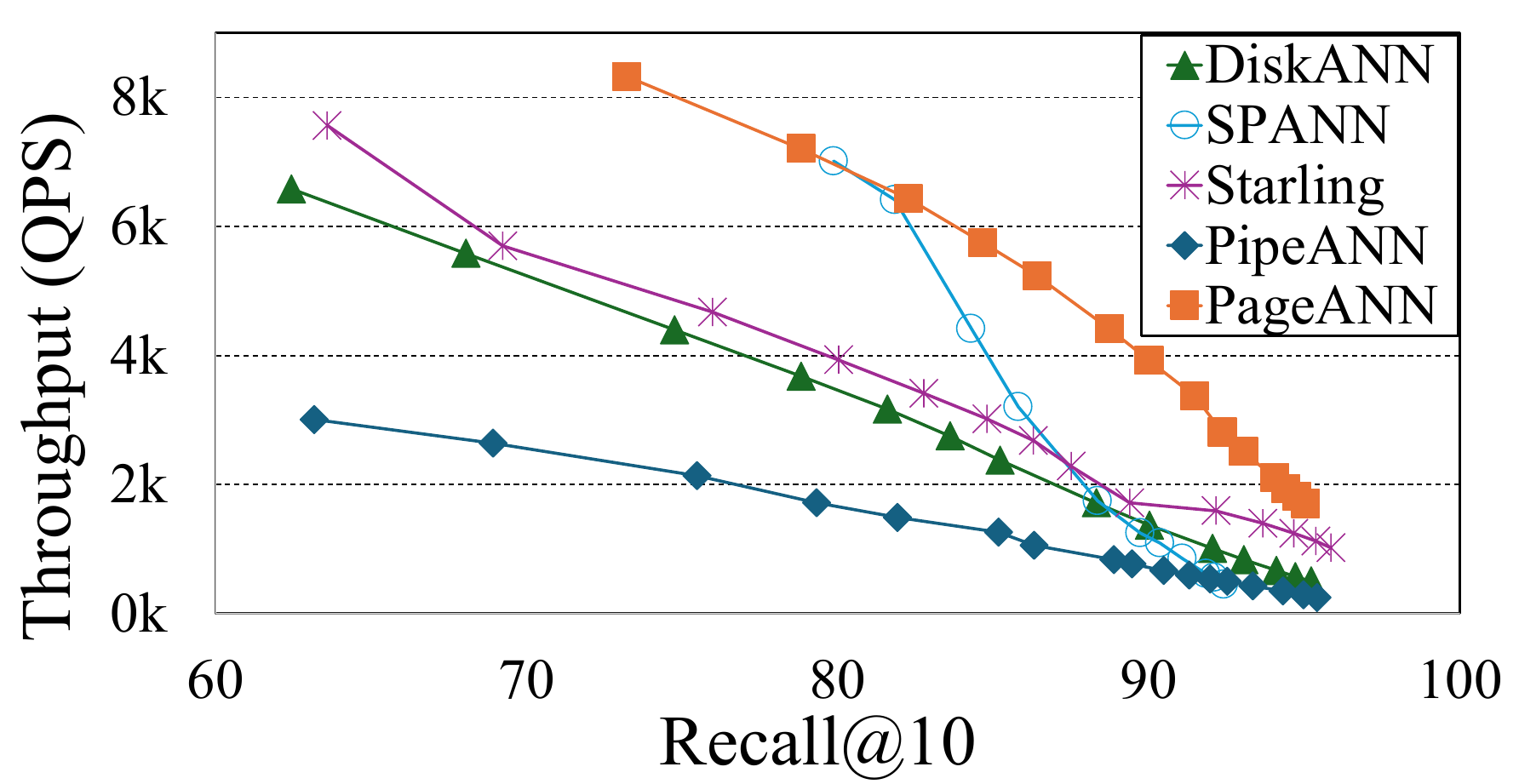}
        \caption{SPACEV100M}
        \label{fig:spacevthroughput_100m}
    \end{subfigure}
    \hfill
    \begin{subfigure}{0.33\textwidth}
        \centering
        \includegraphics[width=\linewidth]{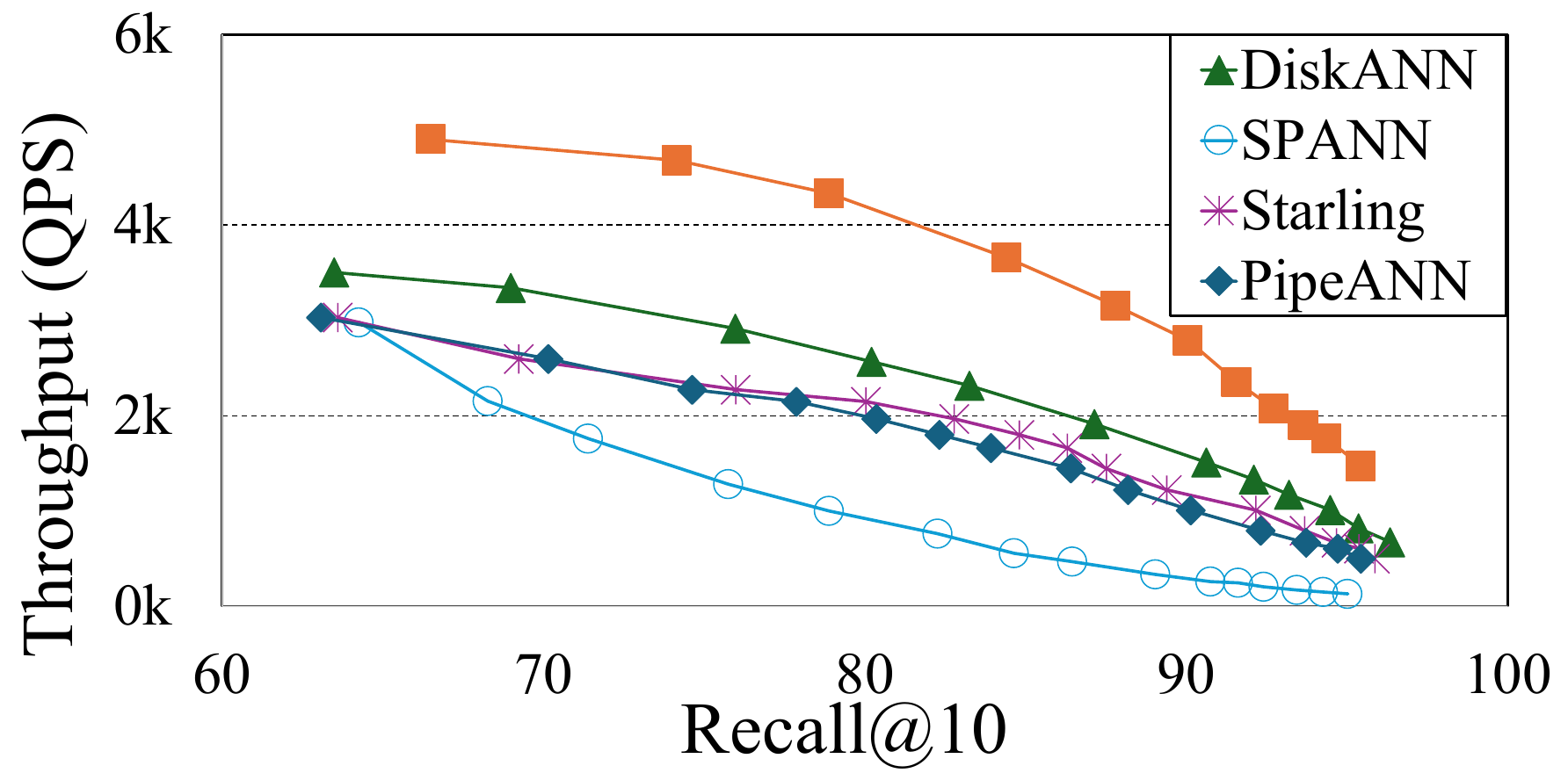}
        \caption{DEEP100M}
        \label{fig:deepthroughput_100m}
    \end{subfigure}
    \caption{The query throughput vs. Recall@10 with the memory ratio of 30\% for all schemes on million-scale datasets.}
    \label{fig:throughput_100m}
\end{figure*}

\noindent\textbf{Memory management for routing and caching.} 
Based on the available memory budget, three main data structures are maintained in memory. First, a contiguous array stores the cached compressed vector values, along with a hash table that maps vector IDs to their corresponding indices in the array. Second, a hash table maps hash codes to the IDs of sampled vectors within their respective subspaces. Third, another contiguous array stores the contents of cached pages, accompanied by a hash table that maps cached page-node IDs to their indices in the array. To reduce the memory overhead of large hash tables, we employ \texttt{tsl::sparse\_map}, which achieves high memory efficiency at the cost of a slight reduction in access speed.


\section{Evaluation}\label{sec:results}


In this section, we evaluates the performance of PageANN against SOTA disk-based ANN schemes under various workloads and memory budgets.

\subsection{Experimental Setup}
\label{sec:exp_setup}


\noindent\textbf{Hardware configuration.}
The experiments are conducted on a workstation running Ubuntu 24.04 under WSL2 with an Intel Core i9-14900K CPU (\SI{4.3}{\giga\hertz}, 24 cores), \SI{64}{\giga\byte} DDR4 RAM RAM, and a 2\,TB NVMe SSD.

\noindent\textbf{Baselines.}
We compare PageANN with four SOTA disk-based ANNS methods including DiskANN~\cite{diskann}, SPANN~\cite{spann}, Starling~\cite{starling}, and PipeANN~\cite{pipeann}. 

\noindent\textbf{Datasets.}
Those schemes are compared on five representative benchmark datasets covering diverse scales and vector dimensionalities: 
SIFT100M~\cite{SIFT}, SPACEV100M~\cite{SPACEV}, DEEP-100M~\cite{DEEP}, 
SIFT1B~\cite{SIFT}, and SPACEV1B~\cite{SPACEV}, 
as shown in Table~\ref{tab:dataset}.

\noindent\textbf{Evaluation metrics.}
We evaluate and report the following key metrics: Recall@10 represents the fraction of the true 10 nearest neighbors retrieved in the top-10 results. Query latency indicates the average per-query processing time in millisecond (ms). Throughput refers to the number of queries processed per second under concurrent load (QPS).

\noindent\textbf{Parameters.}
For a fair comparison, all systems are configured to operate under the same hardware, dataset, and index construction parameters, ensuring identical disk usage and in-memory footprint. Specifically, for DiskANN, Starling, PipeANN, and PageANN, we use the same parameters for average vector degree, vector compression ratio, candidate pool size during construction, and number of vectors per page. During search, the batch size of I/O read requests is fixed at 5, and the number of query threads is fixed at 16 across all datasets. SPANN follows the SPFresh~\cite{xu2023spfresh} configuration, with two modifications:  
(i) the number of vector duplications is tuned to match the disk overhead of other systems, and  
(ii) the head selection ratio is adjusted to ensure a comparable in-memory index size.

\begin{table}[!t]
\centering
\caption{Characteristics of datasets.}
\resizebox{0.99\columnwidth}{!}{
\begin{tabular}{lcccc}
\hline
Dataset      & \# of vectors  & Dimension & Type & \# of queries  \\
\hline
SIFT100M~\cite{SIFT}     & 100M   & 128 & uint8 & 10000  \\
SPACEV100M~\cite{SPACEV}   & 100M   & 100 & int8 & 29316   \\
DEEP100M~\cite{DEEP} & 100M   & 96 & float & 10000       \\
SIFT1B~\cite{SIFT}     & 1B   & 128 & uint8 & 10000  \\
SPACEV1B~\cite{SPACEV}     & 1B   & 100 & int8 & 29316   \\
\hline
\end{tabular}
}
\label{tab:dataset}
\end{table}

\begin{table*}[!t]
\centering
\small
\caption{The comparison of throughput, latency and average I/Os at Recall@10 = 0.9 with the memory ratio of $30\%$.}
\label{tab:million_scale_results}
\resizebox{1\textwidth}{!}{%
\begin{tabular}{lccccccccc}
\toprule
\multirow{2}{*}{Scheme\textbackslash
Dataset} & \multicolumn{3}{c}{SIFT100M} & \multicolumn{3}{c}{SPACE100M} & \multicolumn{3}{c}{DEEP100M} \\
\cmidrule(lr){2-4} \cmidrule(lr){5-7} \cmidrule(lr){8-10}
 & Throughput (QPS) & Latency (ms) & Mean I/Os & Throughput (QPS) & Latency (ms) & Mean I/Os & Throughput (QPS) & Latency (ms) & Mean I/Os \\
\midrule
DiskANN~\cite{diskann}   & 1099.62 & 14.45 & 187.43 & 1369.51 & 11.62 & 141.13 & 1511.31 & 10.50 & 135.32 \\
SPANN~\cite{spann}    & 383.40 & 20.86 & 299.61 & 1089.56 & 14.68 & 96.84 & 257.51 & 62.22 & 296.96 \\
Starling~\cite{starling}  & 1246.94 & 12.75 & 157.79 & 1677.33 & 9.49  & 114.53 & 1008.23 & 15.80 & 189.08 \\
PipeANN~\cite{pipeann}   & 632.16 & 25.16 & 168.51 & 662.73  & 14.19 & 154.47 & 734.33  & 23.98 & 139.52 \\
PageANN (ours)   & \textbf{2749.36} & \textbf{5.78} & \textbf{85.14} & \textbf{3918.03} & \textbf{4.05} & \textbf{59.93} & \textbf{2788.73} & \textbf{5.07} & \textbf{83.94} \\
\bottomrule
\end{tabular}%
}
\end{table*}

\subsection{Overall Performance}
\label{sec:comparison}


\textbf{100M-scale results.}
First, we evaluate the overall performance of all schemes under a memory budget of approximately $30\%$ of the dataset size on three million-scale datasets. Figure~\ref{fig:latency_100m} shows query latency versus recall@10 across the 100-million–scale datasets. PageANN consistently achieves the lowest latency across the entire recall range compared to all baselines. The gap becomes more pronounced at higher recall levels (e.g., Recall@10 = 0.95), where competing methods require substantially larger search lists and consequently incur higher I/O overhead. For example, on SPACEV100M, PageANN achieves an average query latency of 12 ms, whereas Starling and DiskANN exceed 25 ms and 31 ms, respectively.

The throughput results, shown in Figure~\ref{fig:throughput_100m}, lead to a similar conclusion. At Recall@10 = 0.9, PageANN achieves $2.50\times$ higher throughput than DiskANN on SIFT100M, $2.21\times$ higher than Starling, $4.35\times$ higher than PipeANN, and $7\times$ higher than SPANN. Similar margins are observed on SPACEV100M and DEEP100M. As recall increases, competing systems suffer sharp throughput drops due to longer search lists and higher I/O latency. In contrast, PageANN maintains stable performance through page-level traversal and a co-located layout, which together reduce the number of I/O operations.

Table~\ref{tab:million_scale_results} further confirms these findings. At Recall@10 = 0.9, PageANN achieves $46.0\%$ fewer I/O operations, $54.7\%$ lower latency, and more than $85.4\%$ higher throughput than the second-best SOTA scheme. Overall, these results demonstrate that under fair memory budgets ($30\%$ of dataset size), PageANN consistently outperforms competing systems, which either consume more memory (e.g., SPANN, PipeANN) or exhibit significantly higher latency.

\begin{figure}[!t]
    \centering
    \begin{subfigure}{\linewidth}
        \centering
        \includegraphics[width=0.95\linewidth]{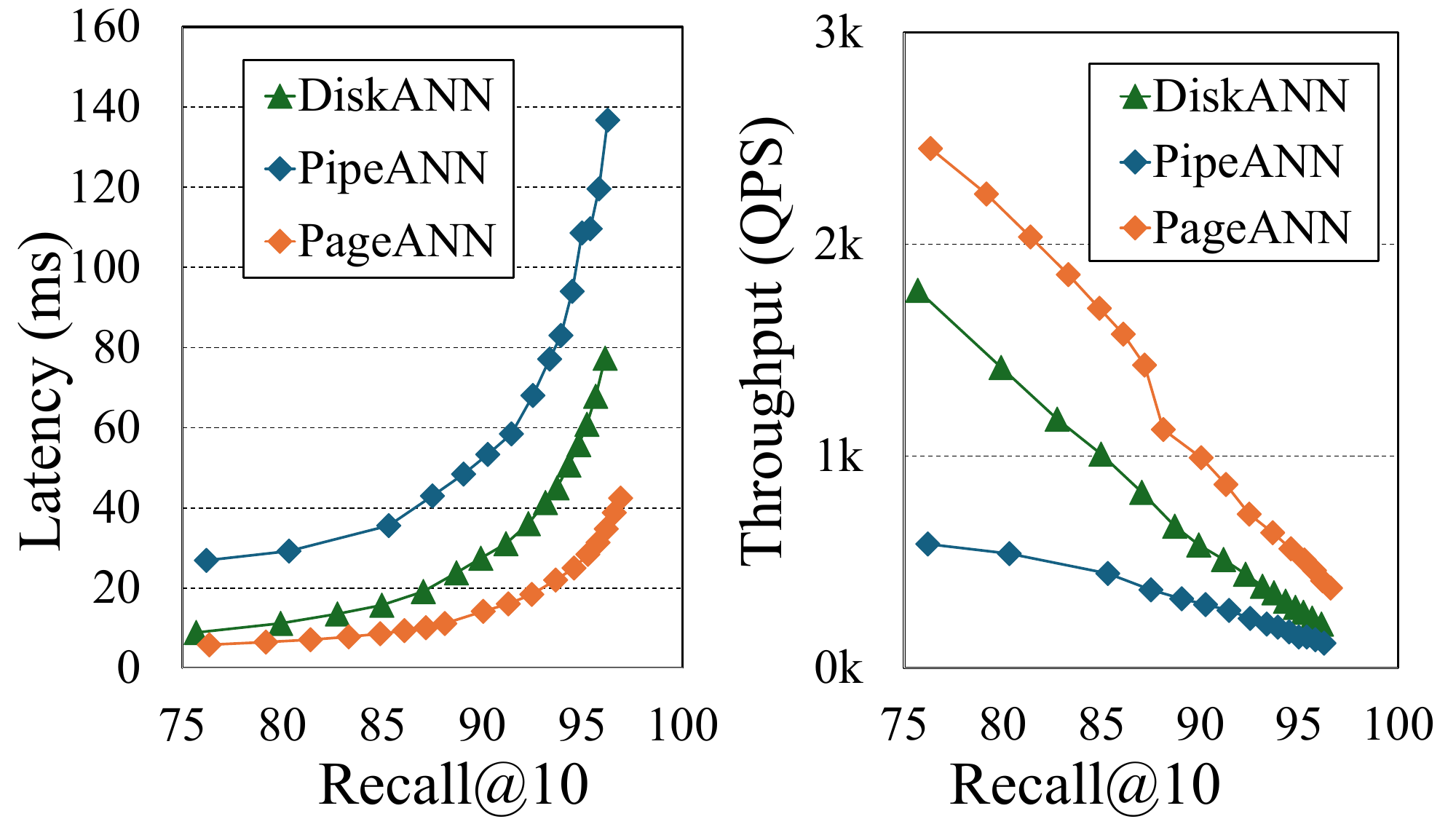}
        \caption{SIFT1B}
        \label{fig:sift1b}
    \end{subfigure}
    \begin{subfigure}{\linewidth}
        \centering
\includegraphics[width=0.95\linewidth]{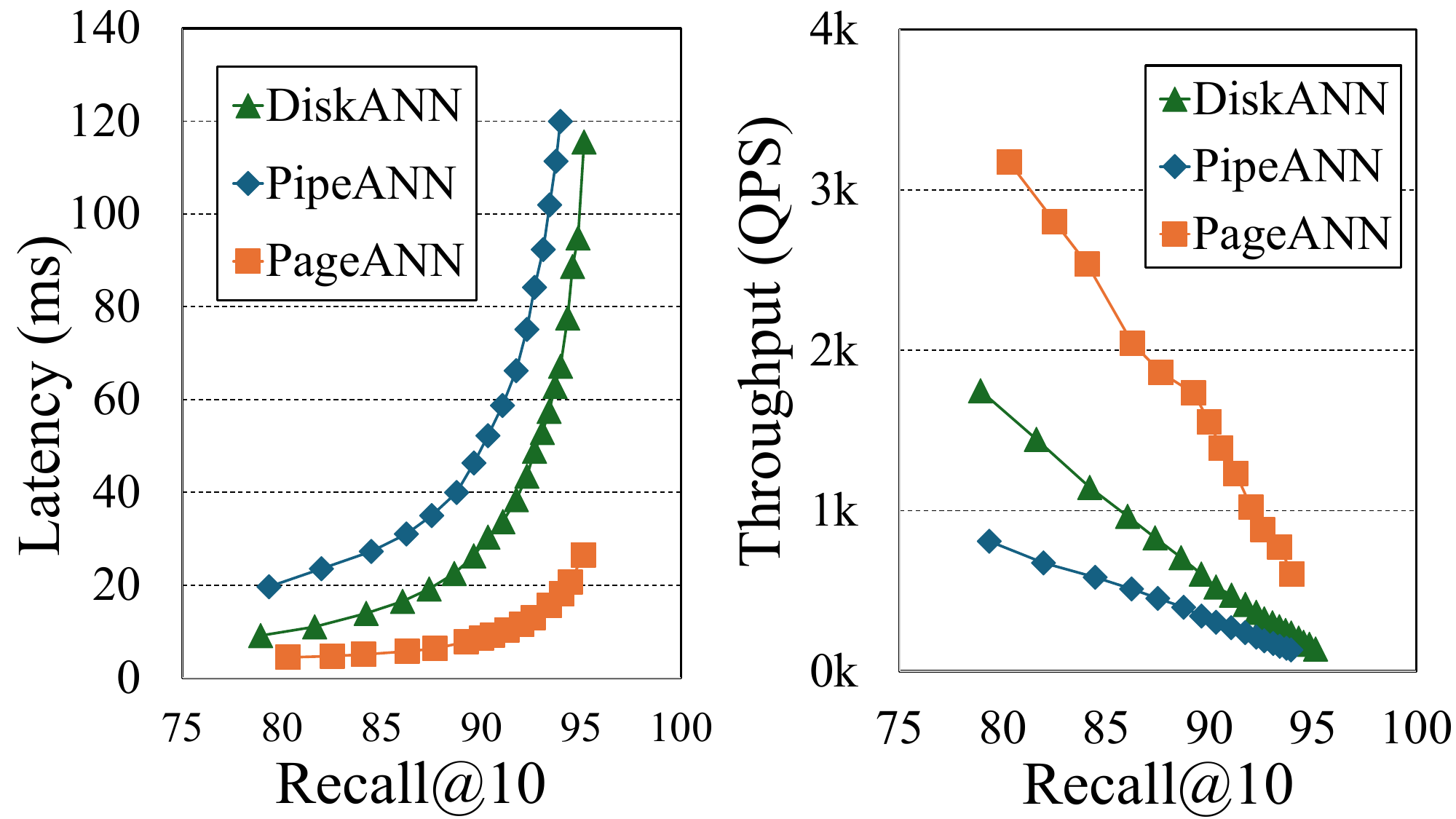}
        \caption{SPACEV1B}
        \label{fig:spacev1b}
    \end{subfigure}

    \caption{Latency and throughput comparison of DiskANN, PipeANN and PageANN on billion-scale datasets (SIFT1B and SPACEV1B).}
    \label{fig:billion_scale_results}
\end{figure}

\textbf{Billion-scale results.} 
We further evaluate PageANN on SIFT1B and SPACEV1B, compared against DiskANN and PipeANN. Other baselines are excluded as their construction or search phase exceeds the memory capacity of our setup. Both PageANN and DiskANN are evaluated with a memory budget of approximately 20\% of the dataset size. For PipeANN, we follow the setup provided in its open-source Github repository, which requires a memory ratio greater than 20\%. As shown in Figure~\ref{fig:billion_scale_results}, PageANN consistently achieves substantially higher throughput and lower latency across the entire recall spectrum. 
On SIFT1B (Figure~\ref{fig:sift1b}) and SPACEV1B (Figure~\ref{fig:spacev1b}), at Recall@10 = 0.9, PageANN delivers up to $1.9\times$--$3.8\times$ higher throughput, $48\%$--$71\%$ lower latency. 
At Recall@10 = 0.95, the advantage widens: PageANN achieves $2.2\times$--$7.5\times$ higher throughput, $55\%$--$87\%$ lower latency. 
These results demonstrate PageANN’s robustness and scalability on billion-scale datasets, where existing methods incur significant I/O bottlenecks and degraded efficiency.

\subsection{Impact of Available Memory on Performance}
\label{sec:ablation}
We examine the impact of available memory (expressed as the memory ratio, i.e., memory budget normalized by dataset size) on the performance of all schemes.

\begin{figure}[!t]
    \centering
    \includegraphics[width=0.95\linewidth]{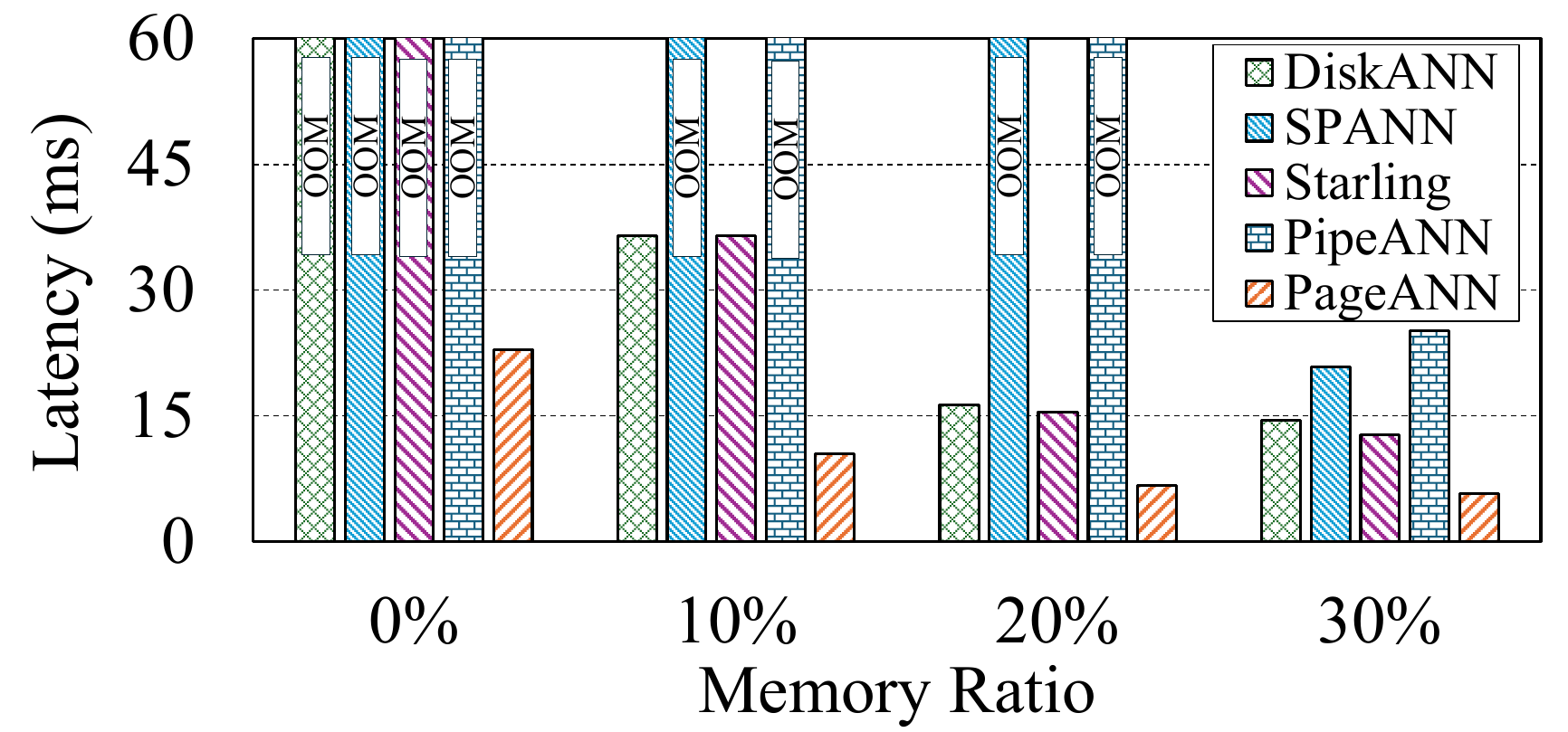}
    \caption{Latency vs. memory ratio from 0\% (around 0.05\%) to 30\% on SIFT100M across different schemes. The label "OOM" means a scheme runs out of memory and cannot operate under the corresponding memory budget.}
    \label{fig:MemoryRatio_vs_Latency}
\end{figure}

\noindent\textbf{Performance sensitivity to memory budget.} Figure~\ref{fig:MemoryRatio_vs_Latency} examines how performance responds as varying memory budget from around $0\%$ to $30\%$ of the dataset size (i.e., memory ratio). PageANN consistently outperforms all baselines across datasets under different memory ratios. PageANN sustains high throughput and low latency across all settings: relative to the $30\%$ configuration, throughput decreases by only $8.7\%$ at $20\%$ memory ratio and $15.2\%$ at $10\%$ memory ratio, with latency increasing only marginally. In contrast, most baselines exhibit sharp performance degradation as memory decreases. DiskANN and PipeANN both suffer significant throughput drops to below $20\%$ of their original values, while SPANN fails to execute reliably when the memory budget falls below $30\%$ memory ratio. These results illustrate the high stability of PageANN under diverse memory scenarios.
 
\noindent\textbf{Minimum memory requirement for high recall.}
The minimum memory required by each system to achieve high recall accuracy is also investigated. Table~\ref{tab:min_mem_req} reports the smallest memory footprint needed to reach Recall@10 = 0.9 on SIFT100M. PageANN achieves the target accuracy and latency with only 0.05 GB ($0.05\%$ of the dataset size), whereas competing systems demand orders of magnitude more memory. Remarkably, even at a memory ratio of around $0\%$, PageANN delivers substantially better performance than state-of-the-art baselines running with a $10\%$ memory ratio.

\begin{table}[!t]
\centering
\small
\caption{Minimum memory size (GB) on SIFT100M to achieve Recall@10 = 0.9.\color{black}}
\label{tab:min_mem_req}
\resizebox{0.45\textwidth}{!}{%
\begin{tabular}{lccccc}
\toprule
\textbf{Method} & DiskANN & SPANN & Starling & PipeANN & PageANN \\
\midrule
\textbf{Min. Size} & 1.2 & 3.2 & 1.2 & 5.4 & \textbf{0.05} \\
\bottomrule
\end{tabular}
}
\end{table}

\begin{figure}[!t]
    \centering
    \begin{subfigure}{0.48\columnwidth}
        \centering
        \includegraphics[width=\linewidth]{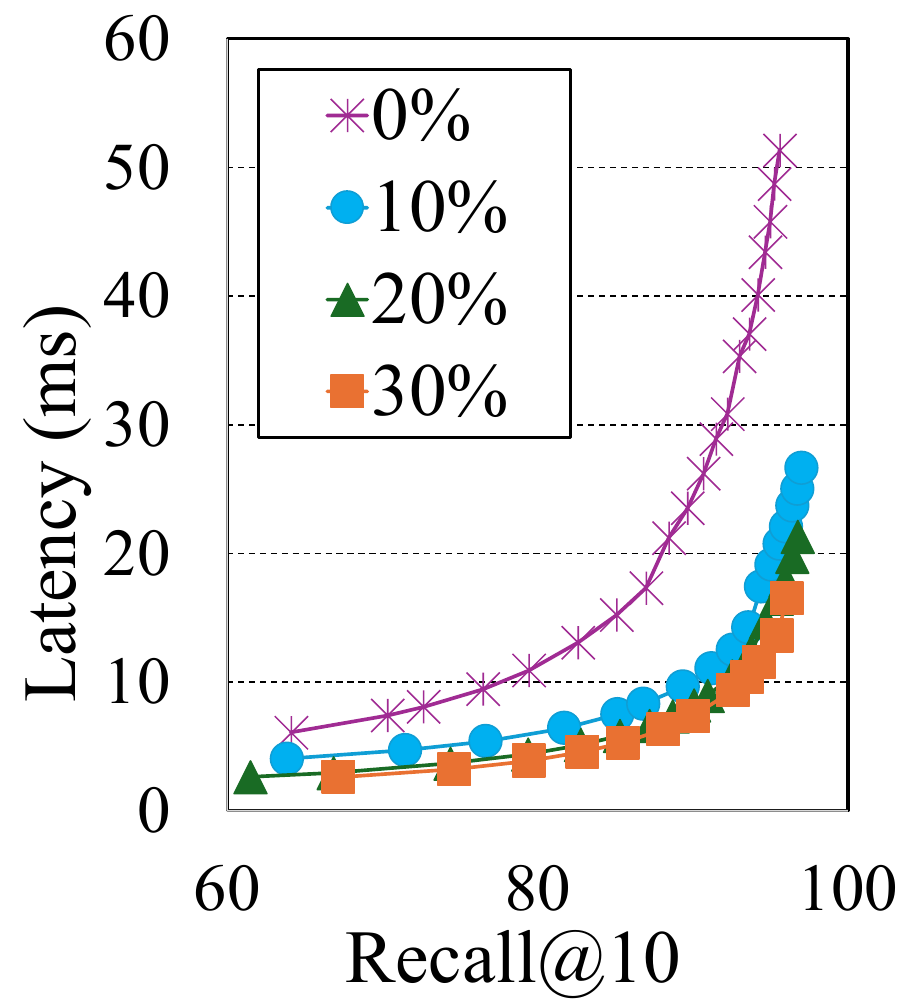}
        \caption{Latency}
        \label{fig:ablation_latency}
    \end{subfigure}
    \begin{subfigure}{0.48\columnwidth}
        \centering
        \includegraphics[width=\linewidth]{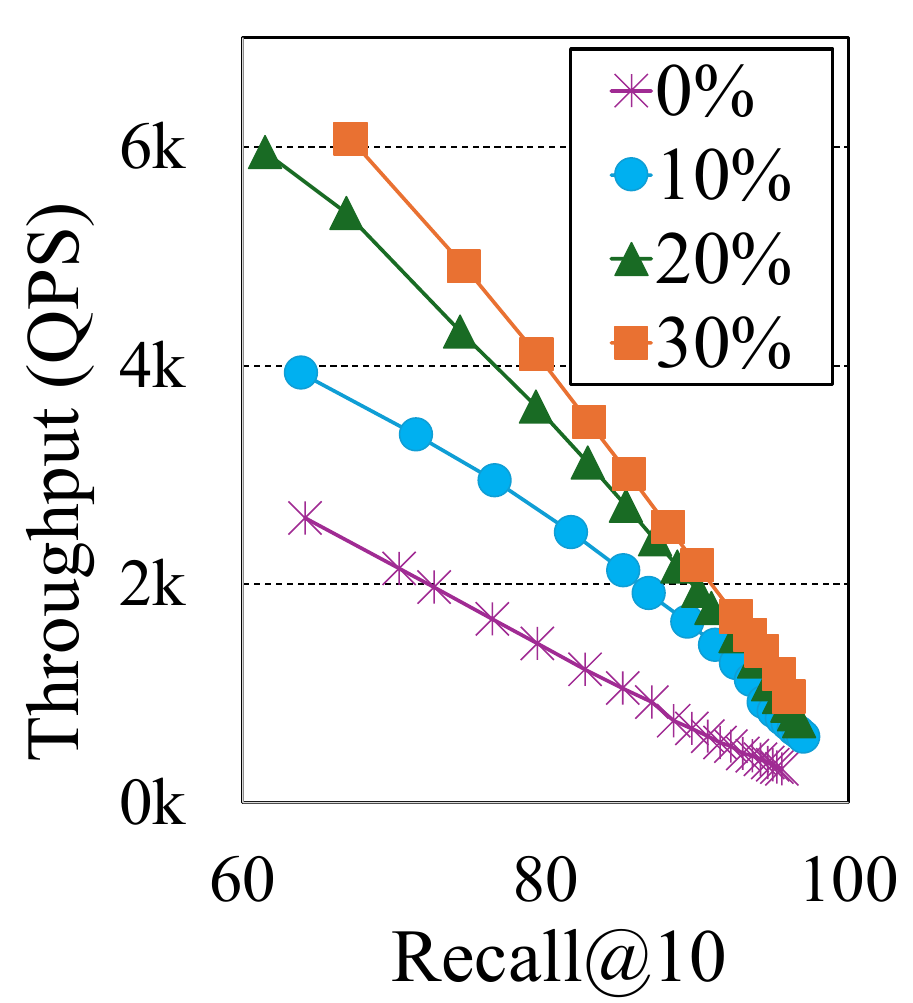}
        \caption{Throughput}
        \label{fig:ablation_throughput}
    \end{subfigure}

    \caption{Latency and throughput of PageANN as varying memory ratio and recall accuracy on SIFT100M.}
    \label{fig:ablation_results}
\end{figure}

\noindent\textbf{Performance of PageANN under varying memory ratio.} As shown in Figure~\ref{fig:ablation_results}, the latency and throughput performance of PageANN improves significantly as the memory budget increases from $0\%$ to $10\%$. This improvement arises because the memory–disk coordination in PageANN allows low-compression vectors to be leveraged even under tight memory constraints. When the budget grows from $10\%$ to $20\%$, the performance gains are more pronounced because all compressed vectors can now fit in memory and each page contains more vectors. As a result, the page-node graph shrinks further and the search path shortens. In addition, the in-memory fast routing procedure is applied. By contrast, the performance increase from $20\%$ to $30\%$ is relatively modest, mainly due to the use of additional cached pages.


\subsection{Throughput and Scalability}
\label{sec:scalability}
\begin{figure}[!t]
    \centering
    \begin{subfigure}{0.48\columnwidth}
        \centering
        \includegraphics[width=\linewidth]{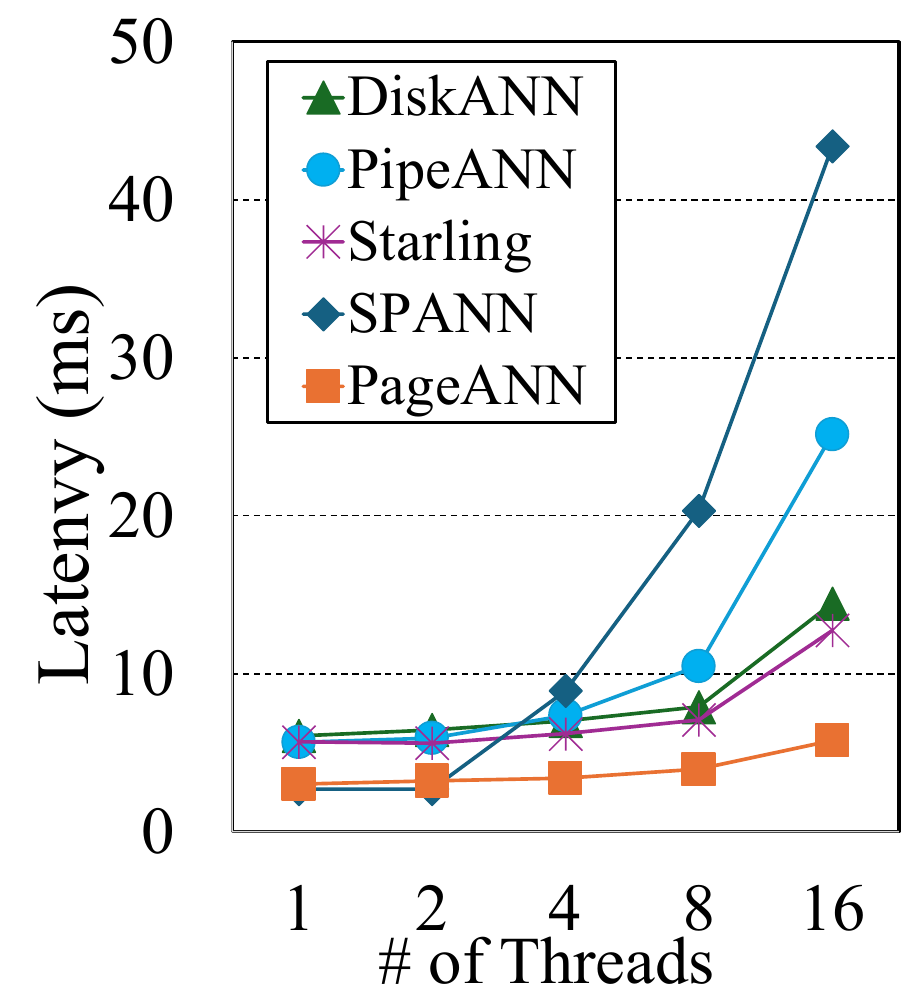}
        \caption{Latency}
        \label{fig:latency_vs_threads}
    \end{subfigure}
    \begin{subfigure}{0.48\columnwidth}
        \centering
        \includegraphics[width=\linewidth]{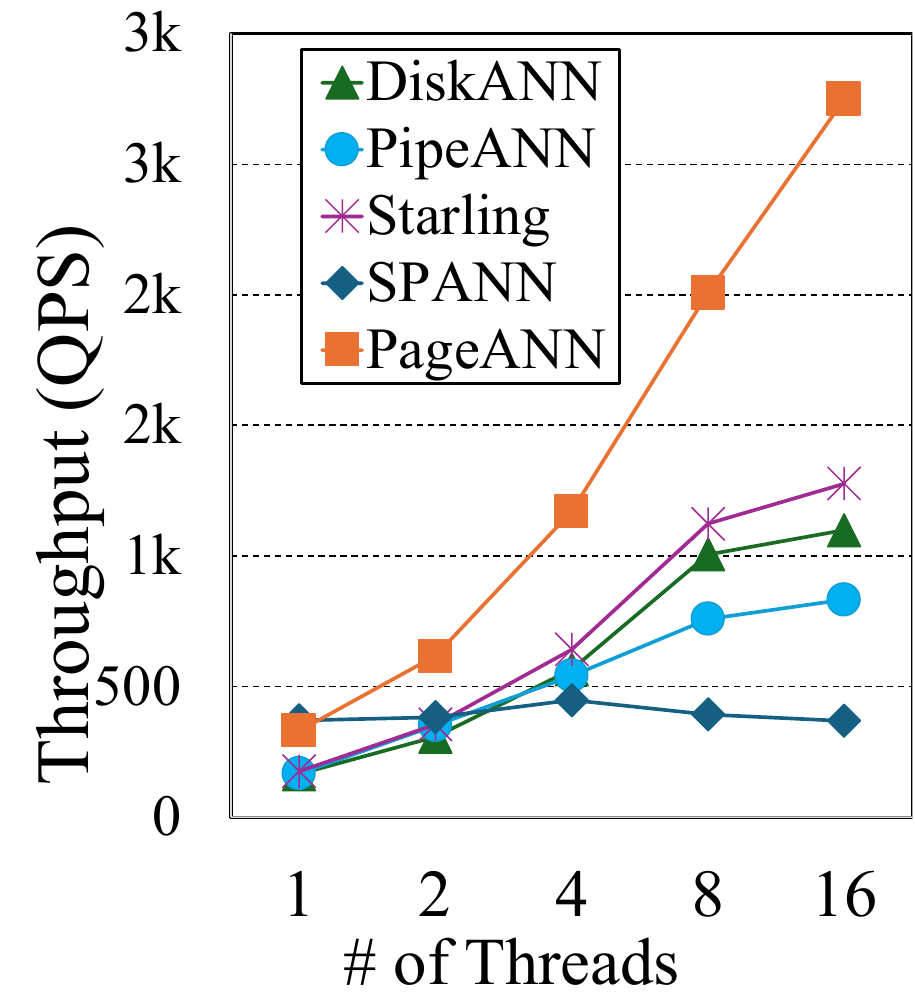}
        \caption{Throughput}
        \label{fig:throughput_vs_threads}
    \end{subfigure}
\caption{Latency and throughput comparison as varying the number of query threads on SIFT100M (Recall@10 = 0.9)}
    \label{fig:scalability}
\end{figure}

We evaluate the scalability of PageANN by measuring its throughput and average query latency as the number of concurrent query threads increases from 1 to 16 on the SIFT100M dataset with Recall@10=0.9 as shown in Figure~\ref{fig:scalability}.
As the number of threads increases, PageANN achieves near-linear throughput scaling, with an $8.34\times$ improvement from single-thread to 16-thread execution. At 16 threads, PageANN delivers $150\%$ higher throughput than DiskANN, $230\%$ higher than PipeANN, and $115\%$ higher than Starling. Latency remains low even under high concurrency, increasing by less than $92\%$ from 1 to 16 threads, in contrast to DiskANN where latency more than triples and PipeANN where it increases over fivefold.
Overall, PageANN achieves both high throughput and low latency across varying concurrency levels, demonstrating strong scalability for large-scale ANN search.

\subsection{Overhead Analysis}
\label{sec:overhead_analysis} 
We evaluate the offline graph construction time and the online CPU utilization during query processing for each method across the three representative datasets. As shown in Table~\ref{tab:graph_build_time}, PipeANN exhibits slightly higher graph contruction time compared to the fastest schemes, due to the additional page-node graph construction. Beyond build time, PageANN generally show higher CPU usage, as it adopts a pipeline design and fully utilizes all data in each loaded page during computation.
In summary, while PageANN introduces a modest computational and pre-processing overhead compared to the fastest scheme, this overhead is acceptable given the substantial performance gains it delivers.

\begin{table}[!t]
\centering
\small
\caption{Graph construction time (hours) during pre-processing and CPU utilization (\%) during query process.}
\label{tab:graph_build_time}
\resizebox{0.95\columnwidth}{!}{%
\begin{tabular}{l ccc ccc ccc}
\toprule
Method   & \multicolumn{2}{c}{SIFT100M} & \multicolumn{2}{c}{SPACE100M} & \multicolumn{2}{c}{DEEP100M} \\
\cmidrule(lr){2-3} \cmidrule(lr){4-5} \cmidrule(lr){6-7}
         & Graph & CPU &Graph & CPU & Graph & CPU \\
\midrule
DiskANN  &  1.20 & 47\% & 1.26 & 145\% & 2.64 & 171\% \\
Starling &  3.03 & 106\% & 2.37 & 144\% & 7.11 & 152\% \\
PipeANN  & 1.21 & 1548\% & 1.27 & 1026\% & 2.65 & 1053\% \\
PageANN   & 1.68 & 168\% & 1.59 & 334\% & 3.50 & 217\% \\
\bottomrule
\end{tabular}%
}
\end{table}


\section{Related Work}\label{sec:related}
\noindent\textbf{In-memory ANNS approaches.} Research on in-memory ANNS is generally categorized into three types: tree-based~\cite{bentley1975multidimensional, fukunaga2006branch, silpa2008optimised}, hashing-based~\cite{lsh98, datar2004locality, tao2009quality, gan2012locality, huang2015query, liu2014sk, lu2020r2lsh, lu2020vhp, tian2023db}, and graph-based methods~\cite{malkov2014approximate, hnsw16, chen2018sptag, fu2017fast, nsg19, li2019approximate}. Among these, graph-based approaches stand out in large-scale, high-dimensional search by offering both high speed and accuracy. However, as dataset sizes continue to grow, the memory overhead of in-memory graph-based ANNS becomes a significant bottleneck, limiting their scalability and applicability. PageANN addresses these memory and scalability challenges by offloading the search graph and vector values to SSDs and fully leveraging SSD characteristics to maximize search performance.

\noindent\textbf{Distributed ANNS approaches} To address the memory overhead and scalability issues of in-memory graph-based ANNS, many distributed ANNS methods have been proposed. These methods fall into two categories: independent sharding ~\cite{milvus, deng2019pyramid, wei2020analyticdb} and global indexing~\cite{zhi2025towards}. Independent sharding offers a simpler system design but suffers from computational inefficiency. In contrast, global indexing avoids redundant computation but incurs significant network communication overhead. PageANN, on the other hand, fully resolves the scalability challenges of graph-based ANNS with a simple system design, efficient computation, minimal I/O latency, and no communication overhead.

\noindent\textbf{Disk-based ANNS approaches} To effectively alleviate the memory constraints of graph-based ANNS, disk-based approaches offload part or all search-related data from memory to disk. A pioneer work, DiskANN~\cite{diskann}, retains only compressed vector values in memory while storing the original vectors and graph index on SSDs. Building on this, DiskANN++~\cite{diskann++} and Starling~\cite{starling} optimize SSD page layouts and integrate an in-memory routing process. LM-DiskANN \cite{lm-diskann} offloads all compressed values to disk at the cost of significantly increased disk usage, while PipeANN~\cite{pipeann} aligns the search algorithm with SSD characteristics. In contrast, SPANN~\cite{spann} keeps the graph in memory and offloads vector values to SSDs, performing graph traversal entirely in memory and issuing all I/O requests only after traversal completes. However, as discussed earlier, all these methods still suffer from high search latency and scalability limitations. PageANN addresses the scalability issue while greatly reducing search latency and increasing throughput. Another types of disk-based ANN schemes focus on the dynamic ANN system to mainly minimize the insertion and graph construction overhead, which is out of the scope with this paper~\cite{xu2023spfresh, xu2025place, pound2025micronn, zhong2025lsm}.


\noindent\textbf{Specialized hardware ANNS approaches.} Recent methods such as FusionANNS~\cite{tian2025towards}, HM-ANN~\cite{hm-ann}, and SmartANN~\cite{smartann} address memory and I/O bottlenecks by leveraging specialized hardware. FusionANNS utilizes GPUs in collaboration with CPUs to reduce I/O operations, HM-ANN exploits heterogeneous memory (DRAM + persistent memory module) to balance cost and performance, and SmartANN employs computational storage devices (SmartSSDs) to minimize host I/O operations and enable near-data processing. However, their reliance on specialized hardware limits flexibility and increases deployment complexity. Unlike these methods, PageANN does not depend on specialized hardware and can be deployed on any machine with varying memory budgets.



\section{Conclusion}\label{sec:conclusion}
In this work, we presented PageANN, a scalable disk-based ANN framework that addresses the key limitations of existing state-of-the-art solutions through a co-design of graph structure, storage layout, and memory management. By introducing a page-based graph that aligns with SSD I/O granularity, a storage layout that leverages spatial locality and merging, and a dynamic memory management strategy with lightweight routing, PageANN achieves high throughput and scalability across diverse memory budgets. Experimental results on large-scale datasets demonstrate that PageANN delivers 1.85x–10.83x higher throughput and 51.7\%–91.9\% lower latency across datasets and memory budgets, while maintaining the same high recall accuracy.

\section{Acknowledgment}
This work was partially supported by NSF 2204656, 2343863, 2413520 and 2440611. Any opinions, conclusions, or recommendations expressed in this material are those of the authors and do not necessarily reflect the views of the NSF.

\bibliographystyle{ACM-Reference-Format}
\bibliography{PageANN-sigplan}

\end{document}